\documentclass[twoside]{article}

\usepackage[accepted]{aistats2021}
\usepackage[round]{natbib}

\usepackage[utf8]{inputenc} 
\usepackage[T1]{fontenc}    
\usepackage{hyperref}       
\usepackage{url}            
\usepackage{booktabs}       
\usepackage{amsfonts}       
\usepackage{nicefrac}       
\usepackage{microtype}      

\usepackage{amsmath,amsfonts,bm}
\newtheorem{prop}{Proposition}

\def\eqref#1{equation~\ref{#1}}

\def\1{\bm{1}}

\DeclareMathAlphabet{\mathsfit}{\encodingdefault}{\sfdefault}{m}{sl}
\SetMathAlphabet{\mathsfit}{bold}{\encodingdefault}{\sfdefault}{bx}{n}

\newcommand{\Abs}[1]{\left| #1 \right| }

\usepackage{amsmath}
\usepackage{amsfonts}
\usepackage{amssymb}
\usepackage{wrapfig}
\usepackage{subcaption}
\usepackage{multirow}
\usepackage{mathtools} 
\usepackage{verbatim}
\usepackage{anyfontsize}
\usepackage{color}
\usepackage{tikz}
\usepackage{adjustbox}
\usepackage{enumitem}
\usepackage{algorithm}
\usepackage{algorithmic}

\usepackage{paralist, tabularx}
\newcommand{\bftab}{\fontseries{b}\selectfont}

\newcommand{\cz}[1]{{\color{black} #1}}

\newcommand{\rzz}[1]{{\color{black} #1}}
\begin{document}

%

%
\runningauthor{Ruqi Zhang, Yingzhen Li, Christopher De Sa, Sam Devlin, Cheng Zhang}

\twocolumn[

\aistatstitle{Meta-Learning Divergences of Variational Inference}

\aistatsauthor{ Ruqi Zhang \And Yingzhen Li* \And  Christopher De Sa }

\aistatsaddress{ Cornell University \And  Imperial College London \And Cornell University }
\aistatsauthor{ Sam Devlin \And Cheng Zhang }

\aistatsaddress{ Microsoft Research, Cambridge \And Microsoft Research, Cambridge}
]
\begin{abstract}
Variational inference (VI) plays an essential role in approximate Bayesian inference due to its computational efficiency and broad applicability. Crucial to the performance of VI is the selection of the associated divergence measure, as VI approximates the intractable distribution by minimizing this divergence. In this paper we propose a meta-learning algorithm to learn the divergence metric suited for the task of interest, automating the design of VI methods. In addition, we learn the initialization of the variational parameters without additional cost when our method is deployed in the few-shot learning scenarios. We demonstrate our approach outperforms standard VI on Gaussian mixture distribution approximation, Bayesian neural network regression, image generation with variational autoencoders and recommender systems with a partial variational autoencoder. 
\end{abstract}
\section{Introduction}

Approximate inference is a powerful tool for probabilistic modelling of complex data. Among these inference methods, variational inference (VI) \citep{jordan1999introduction,zhang2018advances} approximates the intractable target distribution through optimizing a tractable distribution. This optimization-based inference makes VI computationally efficient,
thus suitable to large-scale models in deep learning, such as Bayesian neural networks~\citep{blundell2015weight} and deep generative models~\citep{kingma2013auto}. The objective function in VI is a divergence which measures the discrepancy between the approximate distribution and the target distribution. As an objective function, this divergence significantly affects the inductive bias of the VI algorithm. By selecting a divergence, we encode our preference to the approximate distribution, such as whether it should be mass-covering or mode-seeking. 
The Kullback-Leibler (KL) divergence is one of the most widely used divergence metrics. However, it has been criticized for under-estimating uncertainty, leading to poor results when uncertainty estimation is essential~\citep{bishop2006pattern,blei2017variational,wang2018variational}. Many alternative divergences have been proposed to alleviate this issue \citep{bamler2017perturbative,csiszar2004information,hernandez2016black,li2016renyi,minka2005divergence,wang2018variational}.

\begin{figure*}[t]
    \centering
    \begin{tabular}{cccc}
    \includegraphics[width=3.9cm]{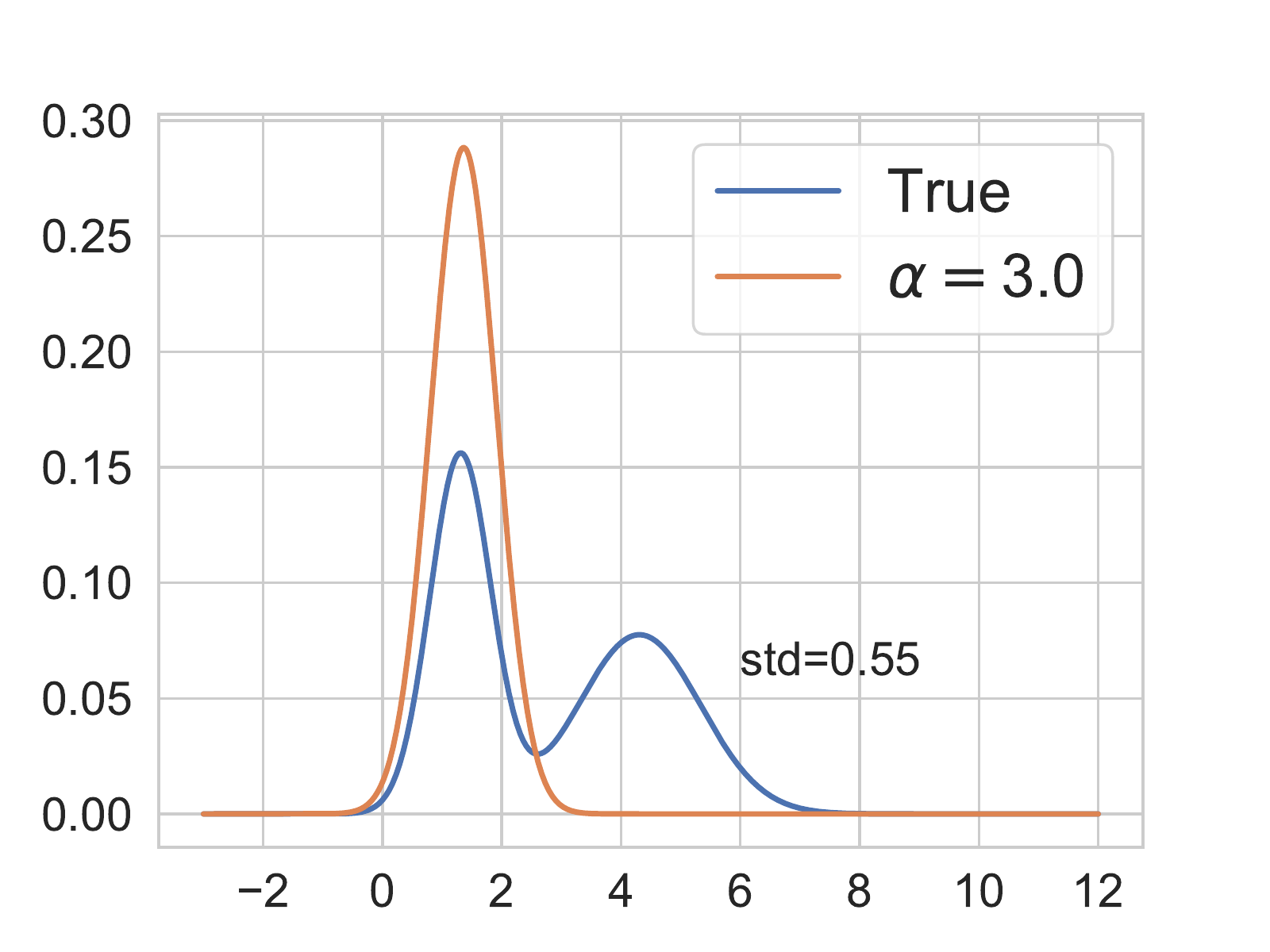}&
    	\includegraphics[width=3.9cm]{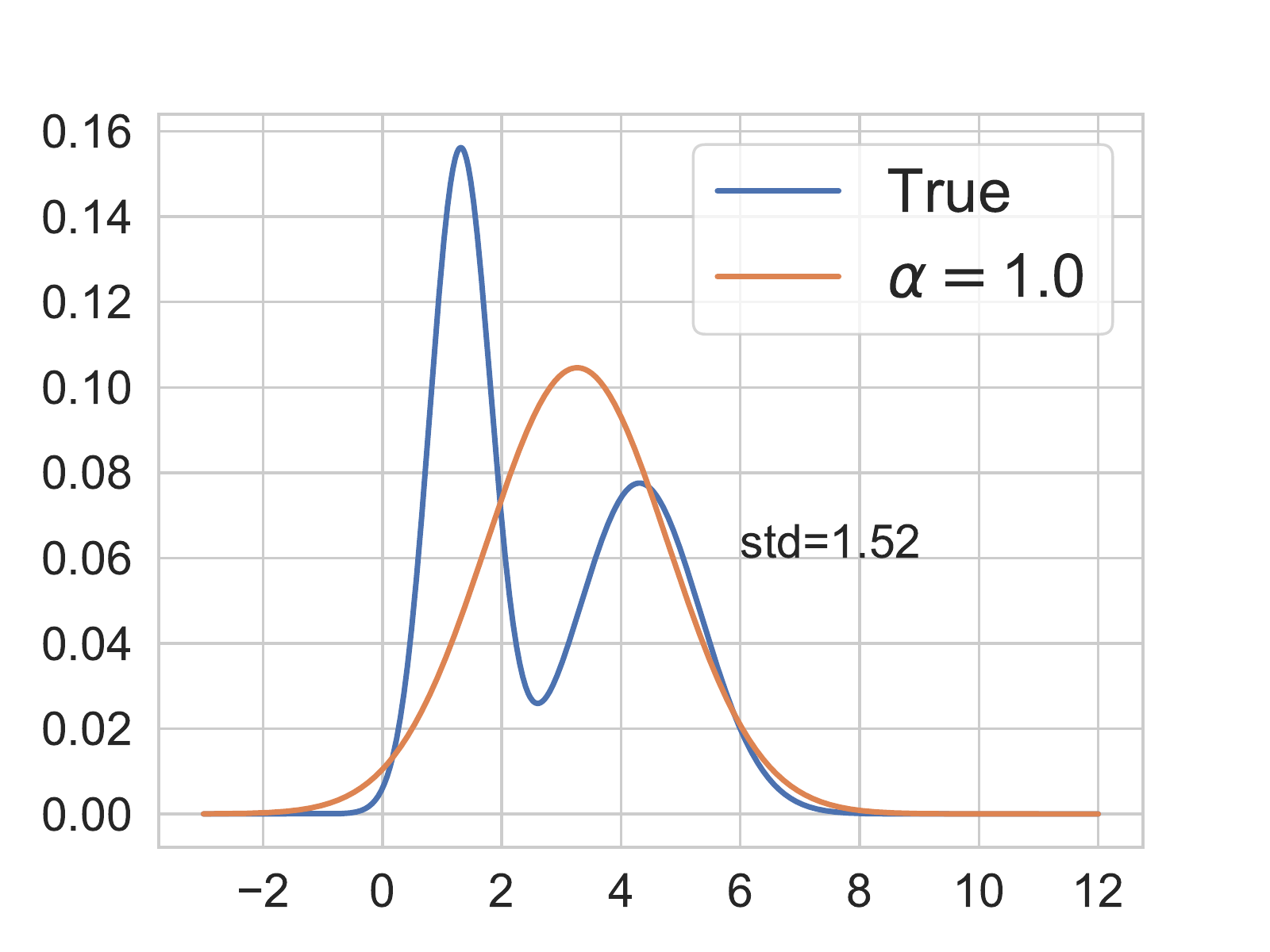} &
    	\includegraphics[width=3.9cm]{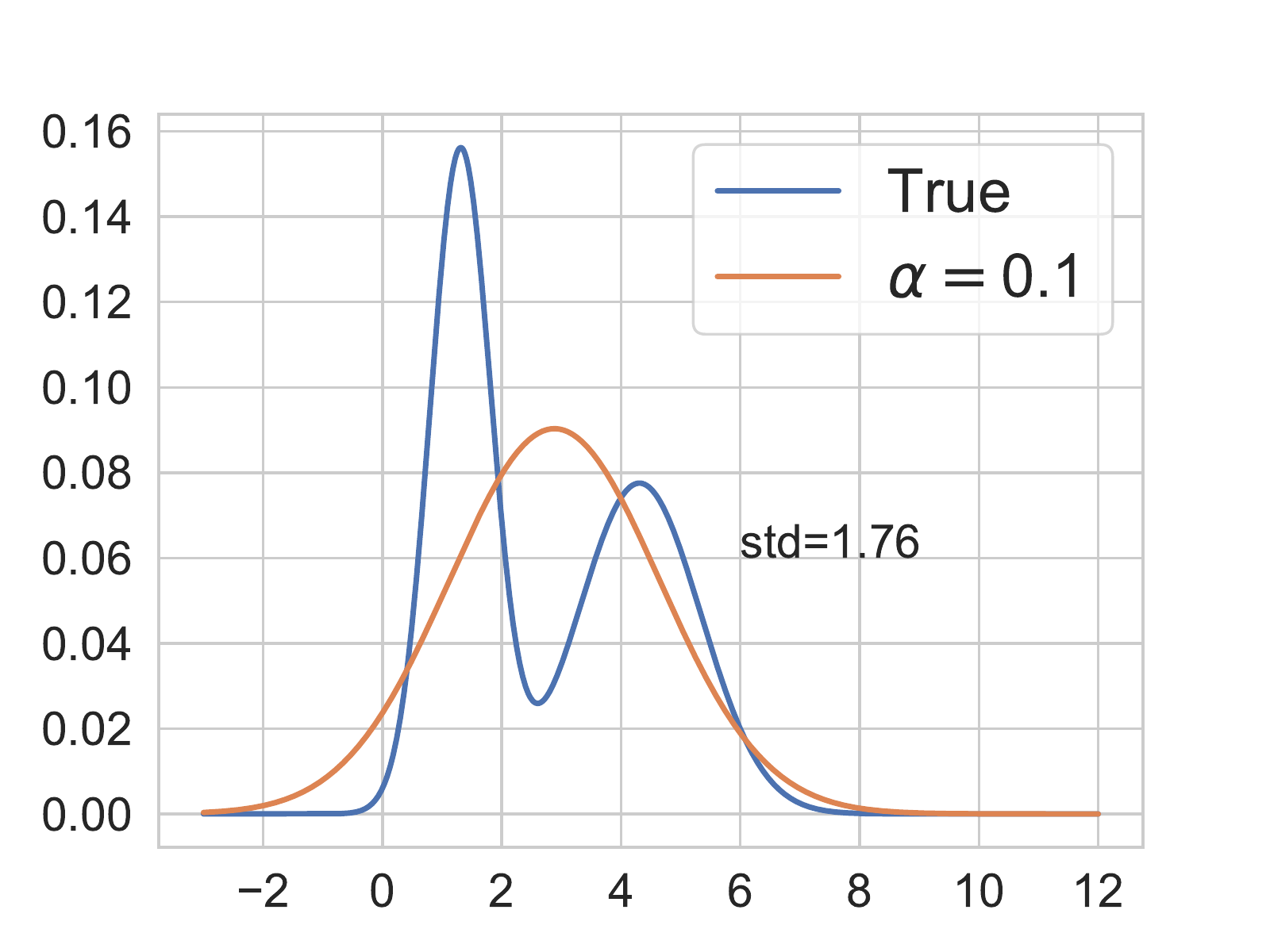} 
    	\\
    	(a) $\alpha=3.0$ & (b) $\alpha=1.0$ & (c) $\alpha = 0.1$
    \end{tabular}
    \caption{An illustration of approximate distributions on a Gaussian mixture by different $\alpha$-divergences (defined in Eq.(\ref{eq:vr-bound})). ``std'' is the standard deviation of the Gaussian approximation.}
    \label{fig:gmm}
\end{figure*}

Although prior work has enriched the divergence family, the optimal divergence metric usually depends on tasks~\citep{minka2005divergence,li2016renyi}. As illustrated by Figure \ref{fig:gmm}, different divergence metrics can lead to very different inference results. Unfortunately, choosing a divergence for a specific task is challenging as it requires a thorough understanding of (i) the shape of the target distribution; (ii) the desirable properties of the approximate distribution; and (iii) the bias-variance trade-off of the variational bound. A crucial question remains to be addressed in order to make VI a success: how can we automatically choose a suitable divergence tailored to specific types of task?

To answer this question, we propose meta-learning divergences of variational inference which utilizes meta-learning, or learning to learn, to refine VI's divergence automatically. In a nutshell, we leverage the fact that various real-world applications consist of many small tasks (e.g.~personalized recommendations for different user groups in recommender systems), and it is important to design a meta-learning algorithm to learn a good inference algorithm for new tasks from previous tasks. We summarize our contributions as follows:

\begin{compactitem}
    \item We develop a general framework for meta-learning variational inference's divergence (Section \ref{sec:metaD}), which chooses the desired divergence objective automatically given a type of tasks. \cz{In this way, we meta-learn the VI algorithm.}
    \vspace{0.2em}
    \item Besides meta-learning the divergence objective, we further meta-learn the parameters for the variational distribution without additional cost (Section \ref{sec:metaDphi}), enabling meta-learning VI in few-shot setting. 
    \vspace{0.1em}
    \item We demonstrate \cz{VI with meta-learned divergences} outperforms standard VI on Gaussian mixture distribution approximation, Bayesian neural network regression, image generation with variational autoencoders, and recommender systems with a partial variational autoencoder (Section \ref{sec:exp}).
\end{compactitem}

\section{Preliminaries}

Consider a dataset $\mathcal{D}=\{x_n\}_{n=1}^N$ and a probabilistic model with parameters $\theta$. Bayesian inference requires computing the posterior over $\theta$ given the dataset $\mathcal{D}$:
$
p(\theta|\mathcal{D}) = p(\mathcal{D}|\theta)p(\theta)/p(\mathcal{D}).
$
The exact posterior is generally intractable, so it needs to be approximated with a tractable posterior $q_{\phi}(\theta) \approx p(\theta | \mathcal{D})$. Typically the approximate posterior $q_{\phi}(\theta)$ is obtained by minimizing a divergence, e.g.~variational inference (VI) often minimizes KL$(q_{\phi}(\theta)\|p(\theta|\mathcal{D}))$. This turns Bayesian inference into an optimization task (divergence minimization). In practice, due to the intractability of $p(\mathcal{D})$, VI alternatively maximizes an equivalent objective called the \emph{variational lower bound}:
\begin{align}\label{eq:variational_lowerbound}
\hspace{-0.9em}
\mathcal{L}_{\text{VI}} = \mathbf{E}_{\theta\sim q_{\phi}}\left[\log\frac{p(\mathcal{D},\theta)}{q_{\phi}(\theta)}\right] = \log p(\mathcal{D}) - \text{KL}(q_{\phi}\|p).
\end{align}

\paragraph{Renyi's $\alpha$-divergence} 
$\alpha$-divergence is a generalization of KL divergence \citep{hernandez2016black,li2016renyi,minka2001expectation}. There are different definitions of $\alpha$-divergence and their equivalences are shown in \citet{cichocki2010families}. Here we focus on Renyi's definition \citep{li2016renyi,renyi1961measures} instead of others \citep{amari2012differential,tsallis1988possible} as it allows our meta-learning framework to be differentiable in $\alpha$ (Section \ref{sec:meta-a}). Renyi's $\alpha$-divergence is defined on $\alpha > 0, \alpha \neq 1$
\begin{align}\label{eq:alpha-divergence}
D_{\alpha}(p\|q) = \frac{1}{\alpha-1}\log\int p(\theta)^{\alpha}q(\theta)^{1 - \alpha}d\theta,
\end{align}
and for $\alpha = 1$ it is defined by continuity: $D_{1}(p\|q) = \lim_{\alpha \rightarrow 1} D_{\alpha}(p\|q) = \text{KL}(p\|q).$
Similar to the variational lower bound, one can maximize the \emph{variational Renyi bound} (VR bound)~\citep{li2016renyi}:
\begin{align}\label{eq:vr-bound}
\mathcal{L}_{\alpha}(q_{\phi}; \mathcal{D}) &= \frac{1}{1-\alpha}\log\mathbf{E}_{\theta\sim q_{\phi}}\left[\left(\frac{p(\theta, \mathcal{D})}{q_{\phi}(\theta)}\right)^{1-\alpha}\right] \\
&= \log p(\mathcal{D}) - D_{\alpha}(q_{\phi}\|p).\nonumber
\end{align}
The expectation is usually computed by Monte Carlo (MC) approximation. To allow gradient backpropagation, the VR bound uses the reparameterization trick \citep{kingma2013auto,salimans2013fixed}, where sampling $\theta \sim q_{\phi}(\theta)$ is conducted by first sampling $\epsilon \sim p(\epsilon)$ from a simple distribution independent with the variational distribution (e.g.~Gaussian) then parameterizing $\theta=r_{\phi}(\epsilon)$. It follows that the gradient of the VR bound w.r.t. the variational parameter $\phi$ after MC approximation with $K$ particles is
\begin{align}\label{eq:alpha-grad} 
   \nabla_{\phi} L_{\alpha}(q_{\phi};x) = \sum_{k=1}^K\left[w_{\alpha, k} \nabla_{\phi}\log\frac{p(r_{\phi}(\epsilon_k), x)}{q(r_{\phi}(\epsilon_k))}\right], 
\end{align}

where {\scriptsize $w_{\alpha, k} = \left(\frac{p(r_{\phi}(\epsilon_k), x)}{q(r_{\phi}(\epsilon_k))}\right)^{1 - \alpha}\bigg/\sum_{k=1}^K\left[\left(\frac{p(r_{\phi}(\epsilon_k), x)}{q(r_{\phi}(\epsilon_k))}\right)^{1 - \alpha}\right]$.} When $\alpha=1$ the weights $w_{\alpha, k} = 1/K$ and the gradient Eq.(\ref{eq:alpha-grad}) becomes an unbiased estimate of the gradient of the variational lower bound Eq.(\ref{eq:variational_lowerbound}).

As shown in Figure \ref{fig:gmm}, approximate inference with different $\alpha$-divergences results in distinct variational distributions. Prior work \citep{li2016renyi,minka2005divergence} also showed the optimal $\alpha$-divergence varies for different tasks and datasets, and in practice it is difficult to choose an optimal $\alpha$-divergence \emph{a priori}.

\paragraph{$f$-divergence}
$f$-divergence defines a more general family of divergences \citep{csiszar2004information, minka2005divergence}. Given a twice differentiable convex function $f : \mathbb{R}_+ \rightarrow \mathbb{R}$, the $f$-divergence is defined as \citep{csiszar2004information}:
\begin{align}\label{eq:f-divergence}
  D_f(p\|q_{\phi}) = \mathbf{E}_{\theta\sim q_{\phi}}\left[f\left(p(\theta)/q_{\phi}(\theta)\right) - f(1)\right].
\end{align}
This family includes KL-divergences in both directions, by taking $f(t)=-\log t$ for KL$(q\|p)$ and $f(t)=t\log t$ for KL$(p\|q)$. It also includes $\alpha$-divergences by setting $f(t) = t^{\alpha}/(\alpha(\alpha-1))$ for $\alpha\in \mathbb{R}\backslash\{0,1\}$. Although the $f$-divergence family is very rich due to its parameterization by an arbitrary twice-differentiable convex function, it requires significant expertise to design a suitable $f$ function for a specific task. Thus the potential of $f$-divergence has not been fully leveraged.

\section{Meta-Learning Divergences of Variational Inference}
\subsection{Problem Set-Up}
The goal of meta-learning VI algorithm is to learn, from a set of tasks, a VI algorithm that produces an approximate distribution with desired properties on new similar tasks. We approach this goal by learning the divergence in use for VI. We formalize the problem setups as follows.

Assume we have a task distribution $p(\mathcal{T})$. Each task $T_i\sim p(\mathcal{T})$ has its own dataset $\mathcal{D}_{T_i}$ and its own probabilistic model $p_{T_i}(\theta_i, \mathcal{D}_{T_i})$. Let $D_{\eta}(\cdot \| \cdot)$ denote a learnable divergence parameterized by $\eta$; then for each task $T_i$ the approximate posterior $q_{\phi_i}(\theta_i)$ is computed by minimizing $D_{\eta}(p_{T_i}(\theta_i | \mathcal{D}_{T_i}) \| q_{\phi_i}(\theta_i))$. In the rest of the paper we write $D_{\eta}(q_{\phi_i}, T_i) =  D_{\eta}(p_{T_i}(\theta_i | \mathcal{D}_{T_i}) \| q_{\phi_i}(\theta_i))$ for brevity. 
To do meta-training, in each step we first sample a minibatch of tasks $T_i, i=1,\ldots,M$ from $p(\mathcal{T})$. Then we define a meta-loss function $\mathcal{J}(q_{\phi_i},T_i)$, and optimize the total meta-loss across all training tasks in the minibatch $\sum_{i=1}^M \mathcal{J}(q_{\phi_i},T_i)$ over the divergence parameter $\eta$. This meta-loss function is designed to evaluate the desired properties of the approximate distribution for these tasks, e.g.~\rzz{negative} log-likelihood. During meta-testing, a new task is sampled from $p(\mathcal{T})$, and the learned divergence $D_{\eta}$ is used to optimize the variational distribution $q_{\phi}$.

We also consider (in Section \ref{sec:metaDphi}) a few-shot learning setup similar to the model-agnostic meta-learning (MAML) framework \citep{finn2017model}. In this case, each task only has a few training data, therefore it is crucial to learn a good model initialization to avoid overfitting and adapt fast on unseen tasks. The goal of meta-learning VI algorithm in this setting is to obtain a divergence as well as an initialization of the variational parameters $\phi$ for unseen tasks. During meta-testing, we will train the model with the learned divergence and the learned initialization of variational parameters on new tasks.

The above two meta-learning settings are practical as demonstrated in many previous works \citep{finn2017model,finn2018probabilistic,gong2018meta,kim2018bayesian}, showing that attaining common knowledge from previous tasks is valuable for future tasks.

\subsection{Meta-Learning Divergences (meta-$D$)}
\label{sec:metaD}
We consider the first setting of learning a divergence. We  assume for now $D_{\eta}$ is given in some parametric form; later on we will provide the details of parameterization of two divergence families ($\alpha$- and $f$-divergence) and show how they fit in this framework.
The general idea is to first optimize the approximate posterior by minimizing the current divergence, then update the divergence using the feedback from the meta-loss. Concretely, for each task $T_i$ we perform $B$ gradient descent steps on the variational parameters $\phi_i$ using VI with the current divergence $D_{\eta}$:
\begin{align}\label{eq:phi-update-d}
    \phi_i \leftarrow \phi_i - \beta \nabla_{\phi_i} D_{\eta}(q_{\phi_i}, T_i).
\end{align}
By doing so the updated variational parameters $\phi_i$ are a function of the divergence parameter $\eta$, which we then update by one-step gradient descent using the meta-loss $\mathcal{J}$: 
\begin{align}
\label{eq:meta_update_eta}
    \eta \leftarrow \eta - \gamma \nabla_{\eta} \frac{1}{M}\sum_i \mathcal{J}(q_{\phi_i}, T_i).
\end{align}
We call this algorithm \emph{meta-$D$} for meta-learning divergences, which is outlined in Algorithm \ref{alg:meta-d}. Our algorithm is different from MAML in that MAML's inner and outer loop losses are designed to be the same, prohibiting it to meta-learn the inner loop loss function which is the divergence in VI. The key insight of our approach is that the updated variational parameters are dependant on the inner loop divergence. This dependency enables meta-$D$ to update the divergence by descending the meta-loss with back-propagation through the variational parameters.

\paragraph{Meta-learning within $\alpha$-divergence family}\label{sec:meta-a}

\begin{figure*}[t]
\begin{minipage}[t]{0.48\linewidth}
\begin{algorithm}[H]
   \caption{Meta-$D$}
\begin{algorithmic}\label{alg:meta-d}
   \STATE {\bfseries Input:} $p(\mathcal{T})$: distribution over tasks; $\beta$, $\gamma$: learning rate hyperparameters; initialize $\eta$
   \LOOP
    \STATE  Sample $M$ tasks $T_i\sim p(\mathcal{T})$
    \FORALL{$T_i$}
    \IF{$\phi_i$ does not exist}
    \STATE initialize $\phi_i$ (can have different architectures)
    \ENDIF
      \STATE Update $\phi_i$ with the current divergence:\\ 
      \FOR{$b=1:B$}
        \STATE $\phi_i \leftarrow \phi_i - \beta \nabla_{\phi_i} D_{\eta}(q_{\phi_i}, T_i)$
      \ENDFOR
   \ENDFOR
   \STATE Update $\eta \leftarrow \eta - \gamma \nabla_{\eta} \frac{1}{M}\sum_i \mathcal{J}(q_{\phi_i}, T_i)$
   \ENDLOOP
   \STATE \textbf{Output:} $\eta$
\end{algorithmic}
\end{algorithm}
\end{minipage}
\hfill
\begin{minipage}[t]{0.48\linewidth}
\begin{algorithm}[H]
   \caption{Meta-$D$\&$\phi$}
\begin{algorithmic}\label{alg:meta-d-phi}
   \STATE {\bfseries Input:} $p(\mathcal{T})$: distribution over tasks; $\beta$, $\gamma$, $\tau$: learning rate hyperparameters
   \STATE Initialize $\phi$, $\eta$
   \LOOP
   \STATE  Sample $M$ tasks $T_i\sim p(\mathcal{T})$
   \FORALL{$T_i$}
      \STATE Update $\phi_i$ with the current divergence:\\ 
      \FOR{$b=1:B$}
        \STATE $\phi_i \leftarrow \phi - \beta \nabla_{\phi} D_{\eta}(q_{\phi}, T_i)$ 
      \ENDFOR
   \ENDFOR
   \STATE Update $\phi \leftarrow \phi - \tau \nabla_{\phi} \frac{1}{M}\sum_i \mathcal{J}(q_{\phi_i}, T_i)$;\\  $\eta \leftarrow \eta - \gamma \nabla_{\eta} \frac{1}{M}\sum_i \mathcal{J}(q_{\phi_i}, T_i)$
   \ENDLOOP
 \STATE \textbf{Output:} $\eta$, $\phi$
\end{algorithmic}
\end{algorithm}
\end{minipage}
\end{figure*}

To make $\alpha$-divergence learnable by the meta-$D$ framework (in this case $\eta = \alpha$), it requires the inner-loop updates (Eq.(\ref{eq:phi-update-d})) to be continuous in $\alpha$. This means a naive solution which relies on automatic differentiation of existing $\alpha$-divergences will fail, due to the fact that these $\alpha$-divergences are not twice differentiable everywhere \citep{li2016renyi,minka2005divergence}. Instead, we propose to manually compute the gradient of Renyi’s $\alpha$-divergence (Eq.(\ref{eq:alpha-grad})) which is continuous in $\alpha \in (0, +\infty)$. Specifically we parameterize $\alpha$-divergence by parameterizing its gradient (Eq.(\ref{eq:alpha-grad})) and set $\nabla_{\phi_i}D_{\eta} = -\nabla_{\phi_i}\mathcal{L}_{\alpha}$ in Algorithm~\ref{alg:meta-d}. \cz{We denote meta-learning a divergence within $\alpha$-divergences family as \emph{meta-$\alpha$}.}

\paragraph{Meta-learning within $f$-divergence family} 
We wish to parameterize the $f$-divergence Eq.(\ref{eq:f-divergence}) by parameterizing the convex function $f$ using a neural network, since neural networks are known to be universal approximators and thus can cover diverse $f$-divergences. However, it is less straightforward to specify the convexity constraint for neural networks. 
Fortunately, Proposition \ref{prop:f-gradient} below indicates that the $f$-divergence and its gradient can be specified through its second derivative $f''$ \citep{wang2018variational}. 

\begin{prop}
If $\nabla_{\theta} \log \left(\frac{p(\theta)}{q_{\phi}(\theta)}\right)$ exists, then by setting $g_f(t) = t^2\cdot f''(t)$, we have (with $\theta = r_{\phi}(\epsilon)$)
\begin{align}\label{eq:f-grad}
    \nabla_{\phi} D_f(p\|q_{\phi})& \nonumber
\\= - \mathbf{E}_{\epsilon\sim p(\epsilon)} \bigg[g_f& \bigg(\frac{p(\theta)}{q_{\phi}(\theta)}\bigg)\nabla_{\phi} r_{\phi}(\epsilon) \nabla_{\theta} \log \bigg(\frac{p(\theta)}{q_{\phi}(\theta)}\bigg)\bigg].
\end{align}
\label{prop:f-gradient}
\end{prop}

Therefore it remains to specify $g$ (or $f''$), and the following Proposition \ref{prop:2} guarantees that using non-negative functions as $g$ is sufficient for parameterizing the $f$-divergence family.
\begin{prop}\label{prop:2}
For any non-negative function $g$ on $\mathbb{R}_+$, there exists a function $f$ such that $g(t) = g_f(t) = t^2\cdot f''(t)$. If $g_f(1)>0$, then $D_f(p\|q_{\phi}) = 0$ implies $p=q_{\phi}$.
\end{prop}
See \citet{wang2018variational} for the proofs. Given these guarantees, we propose to parameterize $f$ implicitly by parameterizing $g(t) = g_f(t)$ which can be any non-negative function. We turn the problem into using a neural network to express a non-negative function that is strictly positive at $t=1$. For convenience, we further restrict the form of the function to be 
\begin{align}
    g(t) = \exp(h_{\eta}(t))\label{eq:nn_h}
\end{align}
where $h_{\eta}(t)$ is a neural network with parameter $\eta$.
This definition of $g$ is strictly positive for all $t$, satisfying the assumption of Proposition \ref{prop:2}. By doing so, the $f$-divergence is now learnable through Algorithm \ref{alg:meta-d}, by computing the gradient $\nabla_{\phi_i} D_{\eta}=\nabla_{\phi_i} D_{f_{\eta}}$ with Eq.~(\ref{eq:f-grad}). 

\rzz{With dataset $\mathcal{D}$, the density ratio in Eq.~(\ref{eq:f-grad}) becomes $\frac{p(\theta | D)}{q_{\phi}(\theta)} = \frac{p(\mathcal{D} | \theta)p(\theta)}{q_{\phi}(\theta)p(\mathcal{D})}$. We estimate $p(\mathcal{D})$ through importance sampling and MC approximation. After doing this, $\frac{p(\theta_k | \mathcal{D})}{q_{\phi}(\theta_k)} = \frac{p(\mathcal{D} | \theta_k)p(\theta_k)}{q_{\phi}(\theta_k)}\bigg/ \frac{1}{K}\sum_{k=1}^K \frac{p(\mathcal{D} | \theta_k)p(\theta_k)}{q_{\phi}(\theta_k)}$ which can be regarded as a self-normalized estimator (see Appendix \ref{sec:compute-f-grad} for details). }

Our method is different from \citet{wang2018variational} in the way that we use deep neural networks parameterization and enable learning the $f$-divergence through standard optimization. We denote meta-learning a divergence within $f$-divergences family as \emph{meta-$f$}.
\subsection{Meta-Learning Divergences and Variational Parameters (meta-$D$\&$\phi$)}
\label{sec:metaDphi}

In addition to learning the divergence objective, we also consider the few-shot setting where fast adaptation of the variational parameters to new tasks is desirable. 
Similar to MAML, the probabilistic models $\{p_{T_i}(\theta_i, \mathcal{D}_{T_i})\}$ share the same architecture, and the goal is to learn an initialization of variational parameters $\phi_i \leftarrow \phi$. On a specific task, $\phi$ is adapted to be $\phi_i$ according to the learnable divergence $D_{\eta}$ (which can be $-\mathcal{L}_{\alpha}$ or $D_{f_{\eta}}$):
\begin{align}\label{eq:phi-update-dphi}
    \phi_i \leftarrow \phi - \beta \nabla_{\phi} D_{\eta}(q_{\phi}, T_i).
\end{align}
The updated $\phi_i$ is a function of both $\eta$ and $\phi$.
For meta-update, besides updating divergence parameter $\eta$ with Eq.(\ref{eq:meta_update_eta}), we also use the same meta-loss to update
\begin{align}\label{eq:global-phi-update}
    \phi \leftarrow \phi - \tau\nabla_{\phi} \frac{1}{M}\sum_i \mathcal{J}(q_{\phi_i}, T_i).
\end{align}
We call this algorithm \emph{meta-$D\&\phi$} which meta-learns both the divergence objective and variational parameters' initialization. It is summarized in Algorithm \ref{alg:meta-d-phi}. Similar to the previous section, the divergence families in consideration are $\alpha$- and $f$-divergence (denoted as meta-$\alpha\&\phi$ and meta-$f\&\phi$ respectively).

\section{Experiments}
\label{sec:exp}

We evaluate the proposed approaches on a variety of tasks.
For the mixture of Gaussians task, we perform distribution approximation (no data) and use different meta-losses to directly demonstrate the ability of meta-$D$ (meta-learning divergences) and meta-$D\&\phi$ (meta-learning divergences and variational parameters) to learn the optimal divergence. For all other experiments, we use \rzz{negative} log-likelihood as the meta-loss. For meta-$D$, we use standard VI (KL divergence) and VI with $\alpha=0.5$ divergence which is a comonly used $\alpha$-divergence \citep{li2015stochastic,wang2018variational} as baselines. For meta-$D\&\phi$, we test it in few-shot setup (i.e. few training data), and compare it to learning $\phi$ only which is obtained by Algorithm~\ref{alg:meta-d-phi} without updating $\eta$. During meta-testing, we test this learned $\phi$ with KL divergence (denoted by VI$\&\phi$). We also include results of VI without learning initialization in the few-shot setup as a reference to show the gain of meta-learning initialization. Unless otherwise specified, we set $B=1$. We discussed the effect of this hyperparameter in Appendix \ref{app:hyperparameter} and put details of experimental setting in Appendix~\ref{app:exp}.

\subsection{Approximate Mixture of Gaussians (MoG)}\label{sec:exp:gmm}

We first verify the ability of our methods on learning good divergences using a 1-d distribution approximation problem. Each task includes approximating a mixture of two Gaussians $p$ by a Gaussian distribution $q_{\phi^*}$ attained from
$
\min_{\phi} D_{\eta}(p\|q_{\phi})
$.
The mixture of Gaussian distribution $p(\theta) = 0.5 \mathcal{N}(\theta; \mu_1, \sigma_1^2) + 0.5 \mathcal{N}(\theta; \mu_2, \sigma_2^2)$ is generated by
\begin{align*}
    \mu_1 \sim \text{Unif}[0, 3],\hspace{1em} \sigma_1 \sim \text{Unif}[0.5, 1.0]; \\
    \mu_2 = \mu_1 + 3,\hspace{1em} \sigma_2 = \sigma_1 * 2.
\end{align*}
Therefore each task has a different target distribution but with similar properties (the same $\mu_2 - \mu_1$ and $\sigma_2/\sigma_1$). As shown in Figure \ref{fig:gmm}, the divergence choice has significant impact on the approximation. 

\begin{table}[t]
\begin{minipage}[t]{0.48 \textwidth}
  \centering
\caption{Meta-$D$ on MoG: learned value of $\alpha$. BO (8 iters) has similar runtime as meta-$\alpha$.}
\label{tab:learned-alpha}
\vspace{-0.7em}
\resizebox{0.8\textwidth}{!}{
\begin{small}
\begin{tabular}{lcccc}
\toprule
Methods &  $\alpha=0.5$& TV \\
\midrule
  meta-$\alpha$ &0.52$\pm$0.01 &0.31$\pm$0.01\\
  BO (8 iters) &0.81$\pm$0.03 & 0.69$\pm$0.08\\
  BO (16 iters) &0.54$\pm$0.07 & 0.32$\pm$0.03\\
\bottomrule
\end{tabular}
\end{small}
}
\vspace{0.7em}
\end{minipage}
\hfill
\begin{minipage}[t]{0.48 \textwidth}
\centering
\caption{Meta-$D$ on MoG: rank of meta-loss over 10 test tasks.}
\label{tab:rank-d}
\vspace{-0.7em}
\resizebox{0.8\textwidth}{!}{
\begin{small}
\begin{tabular}{lcccr}
\toprule
Methods & $\alpha=0.5$ &TV\\
\midrule
meta-$\alpha$ &{\bftab 2.10}$\pm$0.70 & 2.10$\pm$0.30\\
meta-$f$  &{\bftab 2.10}$\pm$1.37 &{\bftab 1.00}$\pm$0.00\\
BO (8 iters) &3.50$\pm$0.67 &4.00$\pm$0.00 \\
BO (16 iters) &2.30$\pm$0.90 & 2.90$\pm$0.30\\
\bottomrule
\end{tabular} 
\end{small}
 }
  \end{minipage}
\end{table}

We test our methods with two types of meta-loss $\mathcal{J}$: $D_{0.5}(q\|p)$ and total variation (TV). 
If $D_{0.5}(q\|p)$ is the metric we care about when evaluating the quality of approximation $q$, then a good divergence will be $D_{0.5}(q\|p)$ itself. This case is to verify our method is able to learn the preferred divergence given a rich enough family $\{ D_{\eta} \}$. In practice, the desired evaluation metric for approximation quality (e.g. log-likelihood) typically does not belong to $\alpha$- or $f$-divergence family; to test this scenario we use the total variation distance (TV) to evaluate the performance of our method when meta-loss is beyond the divergence family.

We first test meta-$D$ (meta-learning the divergences, Algorithm \ref{alg:meta-d}). As a baseline, we treat $\alpha$ as a hyperprameter and use Bayesian optimization (BO) \citep{snoek2012practical} to optimize it. Note that BO is not applicable when the divergence set is $f$-divergence which is parameterized by a neural network, therefore BO is only used as a baseline for meta-$\alpha$.

\begin{figure*}[t]
    \centering
    \begin{tabular}{cccc}
    \hspace{-0mm}
    \includegraphics[width=3.7cm]{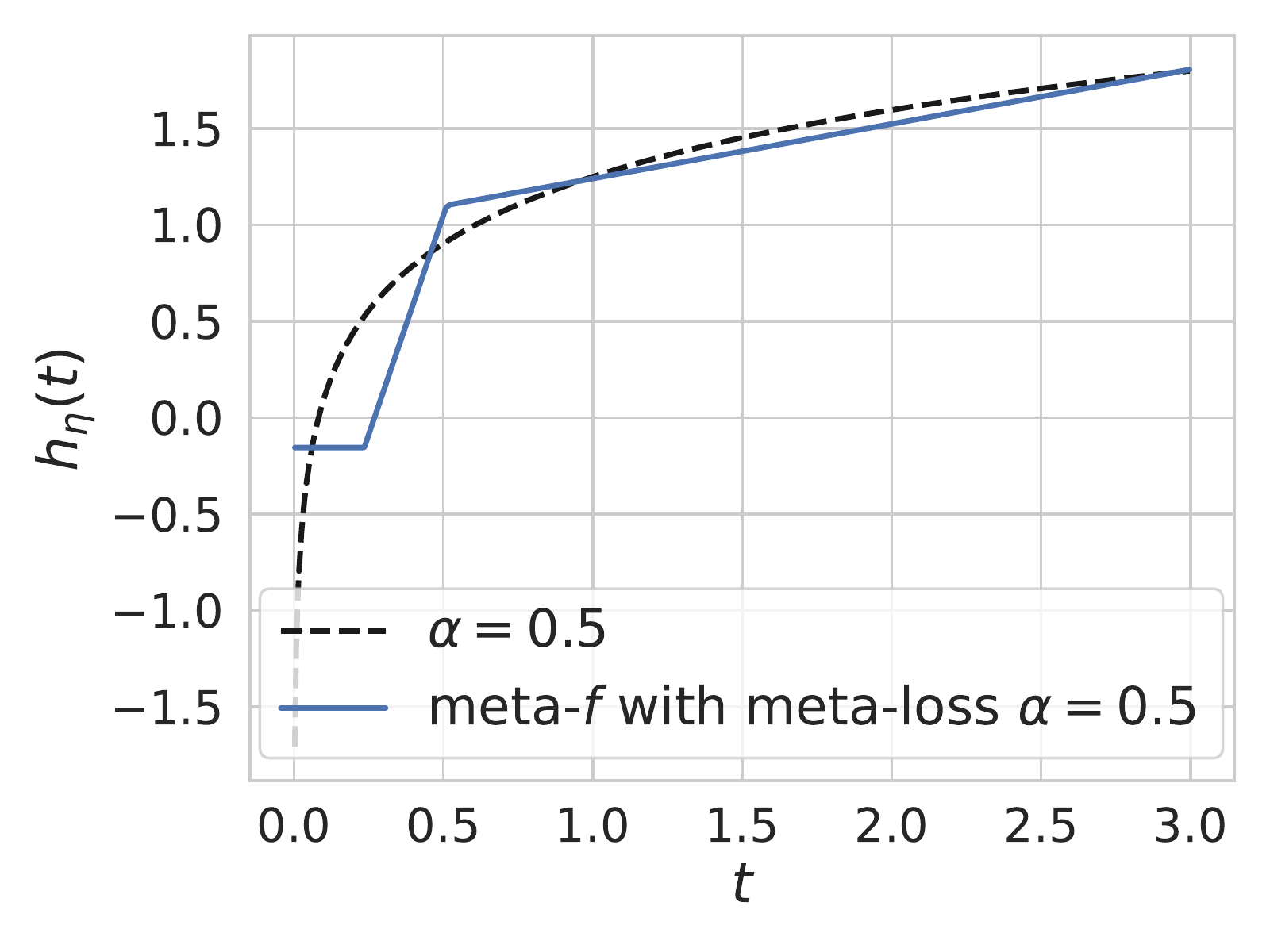} &
    \hspace{-0mm}
    \includegraphics[width=3.7cm]{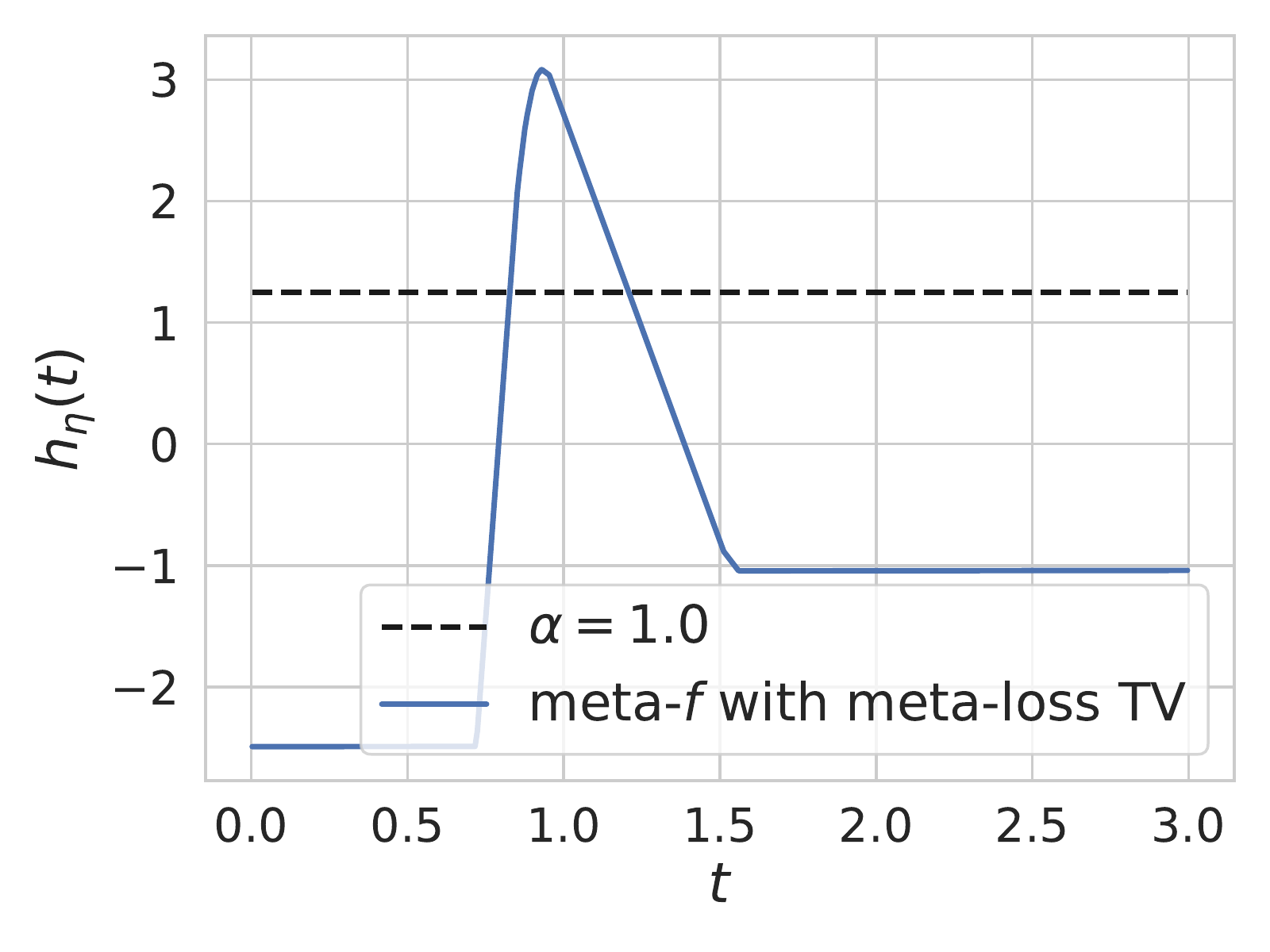}&
    \hspace{-0mm}
    \includegraphics[width=4cm]{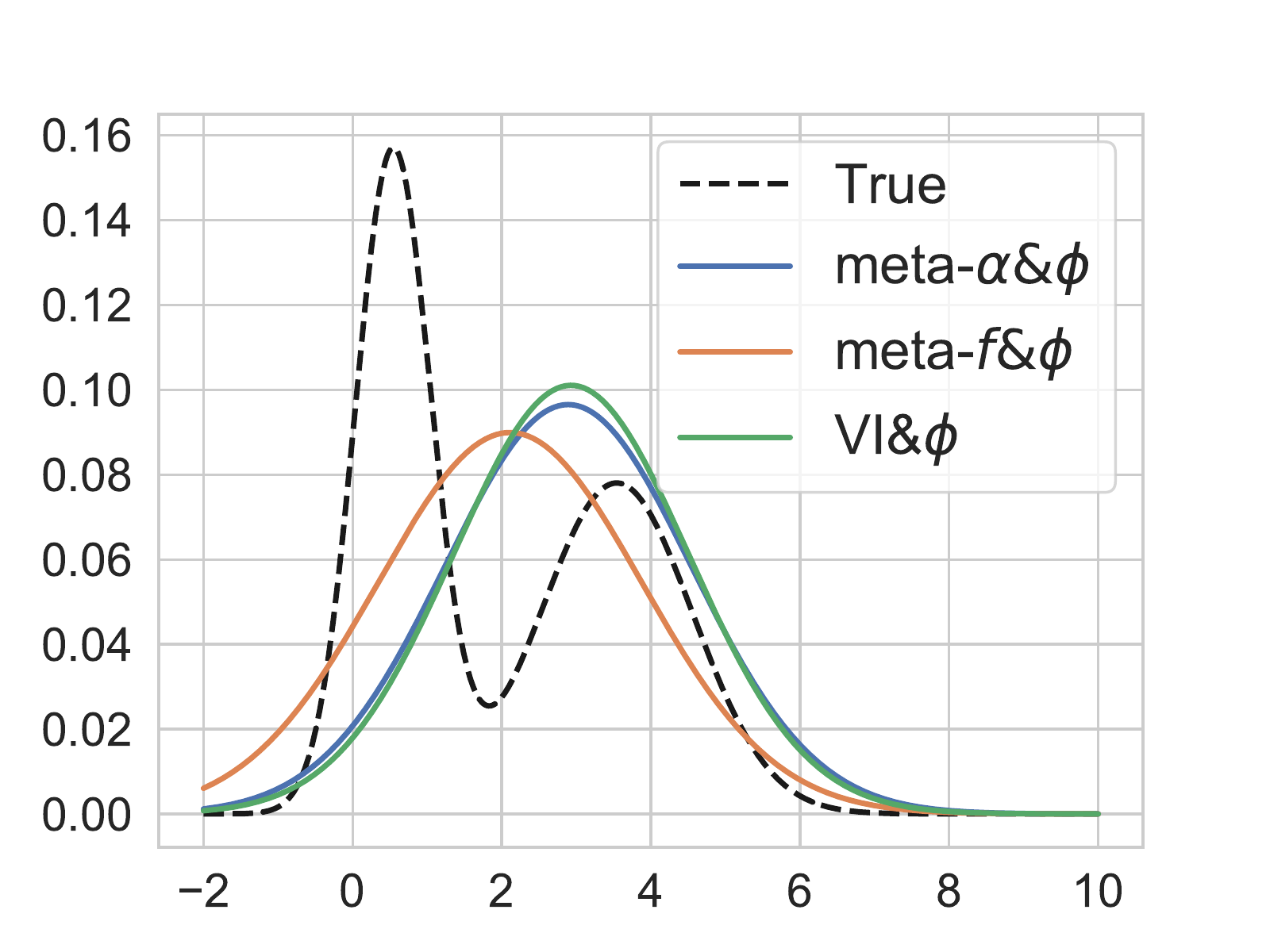}  &
    \hspace{-0mm}
    \includegraphics[width=4cm]{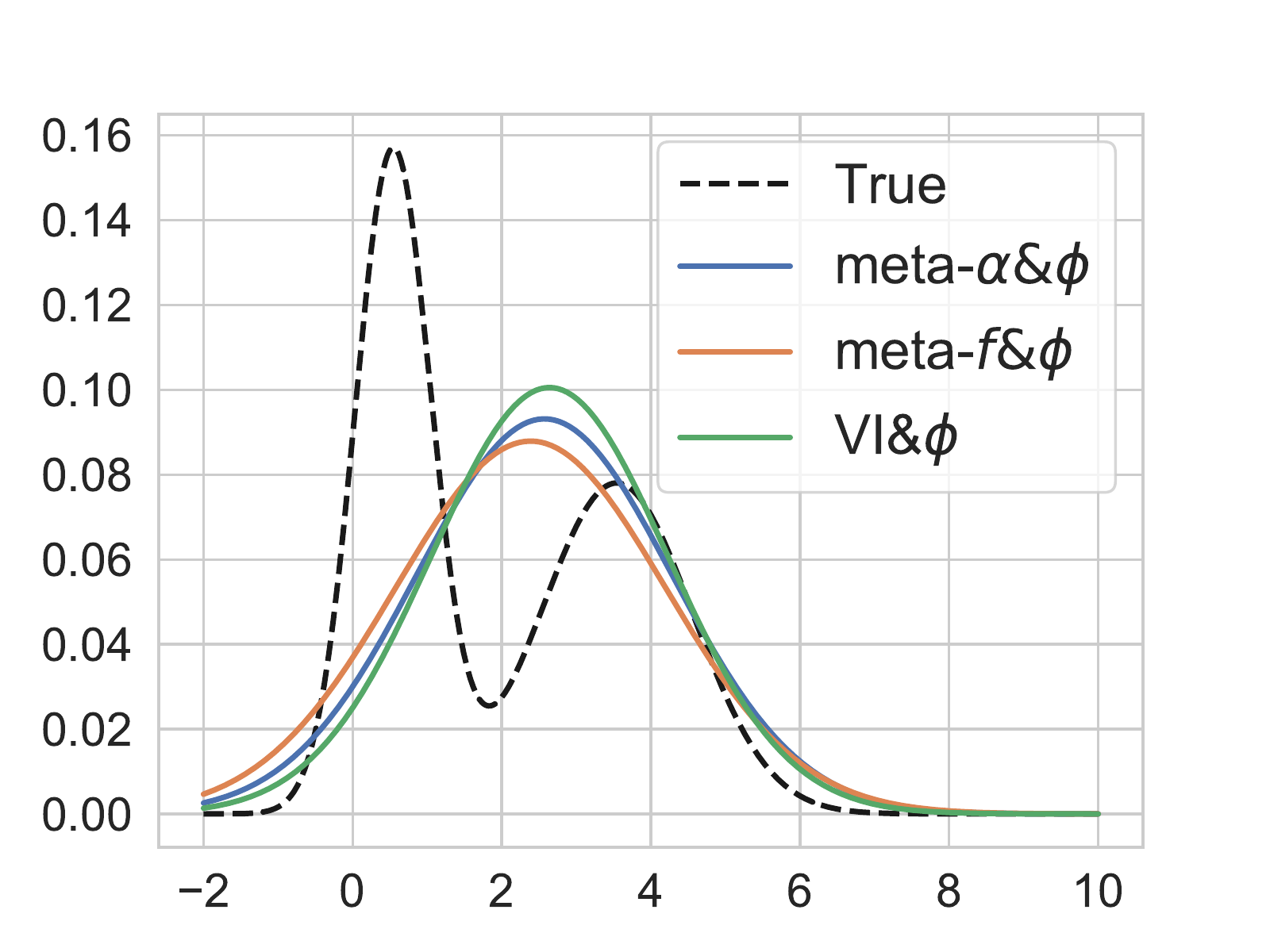} 
    	\\
    	
    	(a) Meta-loss: $\alpha=0.5$&
    	(b) Meta-loss: TV&
    	(c) Meta-loss: $\alpha=0.5$&
    	(d) Meta-loss: TV\\
    \end{tabular}
    \caption{Visualization of (a)-(b) learned $h_{\eta}$ and (c)-(d) approximate distribution after 20 updates. Meta-$f$ refers to meta-learning divergences (Algorithm~\ref{alg:meta-d}) within $f$-divergences. Meta-$\alpha\&\phi$ and Meta-$f\&\phi$ refer to meta-learning divergences and variational parameters (Algorithm~\ref{alg:meta-d-phi}) within $\alpha$- and $f$-divergences respectively. VI$\&\phi$ refers to meta-learning varitional parameters only.
    }
    \label{fig:gmm-dphi}
\end{figure*}

We report the learned values of $\alpha$ from meta-$\alpha$ and BO in Table \ref{tab:learned-alpha}. When the meta-loss is $D_{0.5}$, the learned $\alpha$ from meta-$\alpha$ is very close to 0.5, confirming that our method can pick up a desired divergence. Note that BO is less computationally efficient, as it needs to train a model from scratch every single time when evaluating a new value of $\alpha$, while our method can update $\alpha$ based on the current model.\footnote{We also considered BO in later sections but found it very inefficient (e.g. on the experiment in Section \ref{sec:exp:sin}, BO can only conduct two searches given similar runtime as our methods) thus omitted the results.}  
We test learning $f$-divergence and visualize the learned $h_{\eta}(t)$ (Eq.(\ref{eq:nn_h})) in Figure \ref{fig:gmm-dphi}(a)\&(b). When the meta-loss is $D_{0.5}(q\|p)$, the corresponding $h_{0.5}(t)$ for $D_{0.5}$ is analytical (Appendix~\ref{app:BNNsetting}), and we see from Figure \ref{fig:gmm-dphi}(a) that the learned $h_{\eta}(t) \approx h_{0.5}(t) + 1.25$. This means meta-$D$ has learned the optimal divergence $D_{0.5}$, since $f(t)$ and $af(t)$ define the same divergence for $\forall a > 0$.

When the meta-loss is TV, the optimal divergence is not analytic. Therefore, we instead report the averaged rank of meta-losses on 10 test tasks in Table \ref{tab:rank-d} (see Table \ref{tab:value-metaloss} in Appendix for averaged value of meta-losses). It clearly shows that meta-$\alpha$ and meta-$f$ are superior over BO. Moreover, meta-$f$ outperforms meta-$\alpha$ when the meta-loss is TV. From Figure \ref{fig:gmm-dphi}(b), we can see that the learned $f$-divergence is not inside $\alpha$-divergence, showing the benefit of using a larger divergence family. It also indicates that our $f$-divergence parameterization using a neural network is flexible and can lead to new $f$-divergences that are not used before.

\begin{table}[t]
  \centering
    \caption{Meta-$D$\&$\phi$ on MoG: rank of meta-loss over 10 test tasks. }
  \label{tab:rank-dphi}
\resizebox{0.5\textwidth}{!}{
  \begin{tabular}{ccc|cc}
  \toprule
     Method&{ $\alpha=0.5$}&{TV}&{$\alpha=0.5$}&{TV }\\
       &{ (20 iters)}&{(20 iters)}&{(100 iters)}&{ (100 iters)}\\
              \midrule 
  VI\&$\phi$ &2.70$\pm$0.46 &2.70$\pm$0.46 &2.40$\pm$0.49 &2.50$\pm$0.50\\            
  meta-$\alpha\&\phi$ &2.10$\pm$0.54 &1.80$\pm$0.60 &2.20$\pm$0.75 &{\bftab 1.40}$\pm$0.66 \\
  meta-$f\&\phi$ &{\bftab 1.20}$\pm$0.60 &{\bftab 1.50}$\pm$0.81 &{\bftab 1.40}$\pm$0.80 &2.10$\pm$0.83\\
    \bottomrule 
  \end{tabular}
  }
\end{table}

Next we test meta-$D\&\phi$ (meta-learning divergences and variational parameters, Algorithm~\ref{alg:meta-d-phi}). During training, we perform $B=20$ inner loop gradient updates. The learned $\alpha$ is 0.88 and 0.77 for meta-loss $D_{0.5}$ and TV respectively, which is different from those reported in Table \ref{tab:learned-alpha}. We conjecture that this is related to the learned $\phi$ and $B$ (the horizon length). During meta-testing, we start from the learned $\phi$ and train the variational parameters with the learned divergence for 20 and 100 iterations, corresponding to short and long horizons respectively. Table \ref{tab:rank-dphi} summarizes the rankings. Our methods are better than VI\&$\phi$ (which uses KL and only meta-learns $\phi$) in all cases, demonstrating the benefit of learning a task-specific divergence instead of using the conventional VI for all. 
To further elaborate, we visualize in Figure \ref{fig:gmm-dphi}(c)\&(d) the approximate distributions after 20 steps. The $q$ distributions obtained by meta-$D\&\phi$ tend to fit the MoG more globally (mass-covering), resulting in better meta-losses when compared with VI$\&\phi$. Compared to Algorithm \ref{alg:meta-d}, Algorithm \ref{alg:meta-d-phi} helps shorten the training time on new tasks (100 v.s. 2000 iterations). Notably, meta-$D\&\phi$ is able to provide this initialization along with divergence learning without extra cost.

\subsection{Regression Tasks with Bayesian Neural Networks}\label{sec:exp:sin}

\begin{figure*}[t!]
    \centering
    \begin{tabular}{ccccc}
    \hspace{-5mm}
    \includegraphics[width=3.5cm]{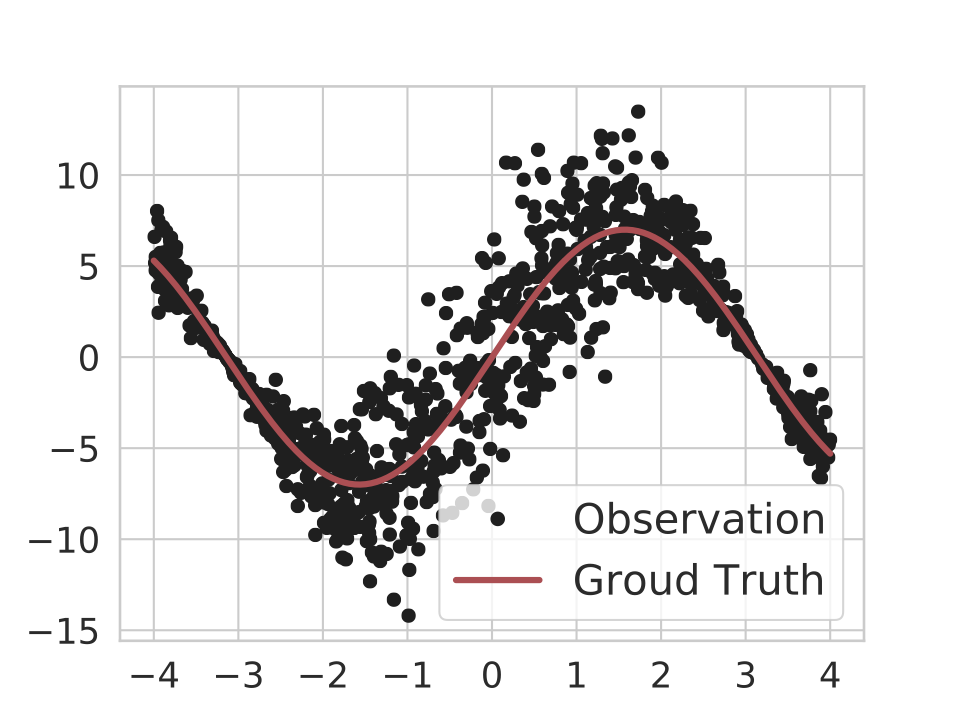}  &
    \hspace{-5mm}
    	\includegraphics[width=3.5cm]{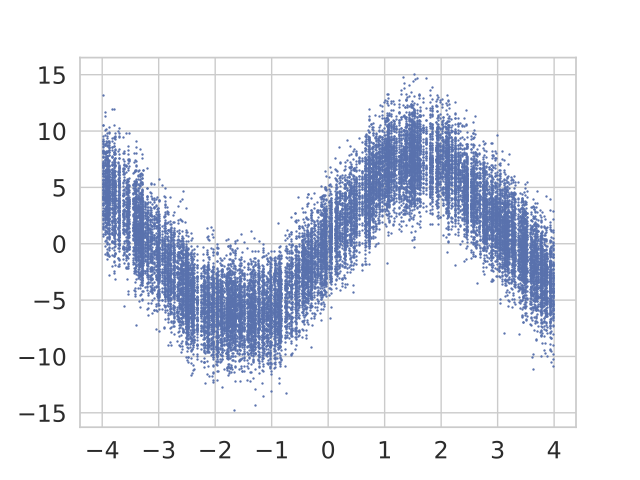} &
    	\hspace{-5mm}
    	\includegraphics[width=3.5cm]{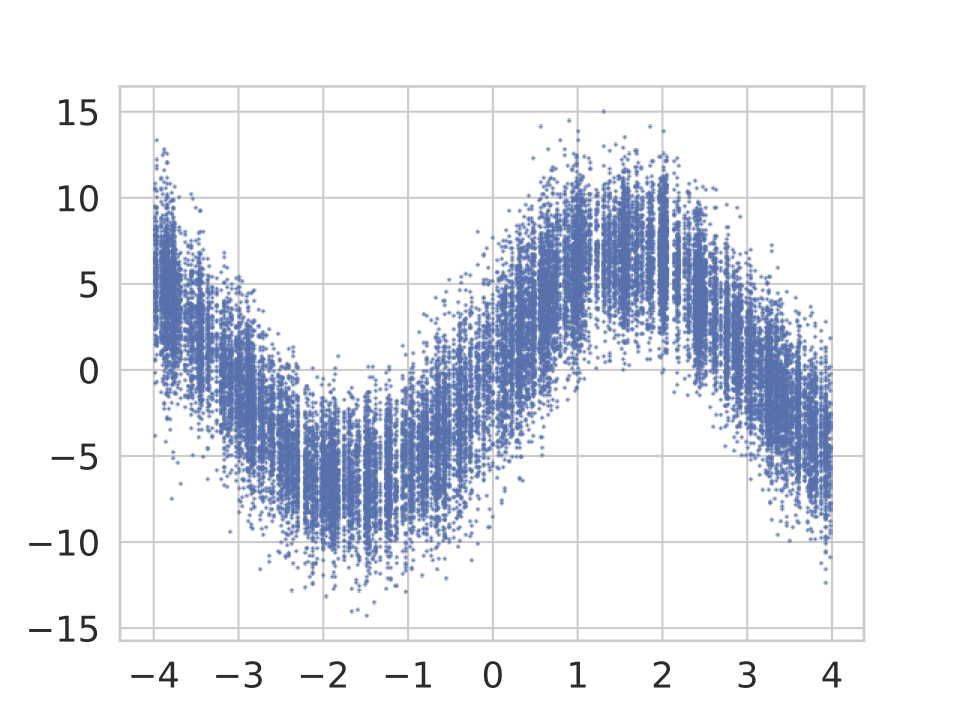} &
    	\hspace{-5mm}
    	\includegraphics[width=3.5cm]{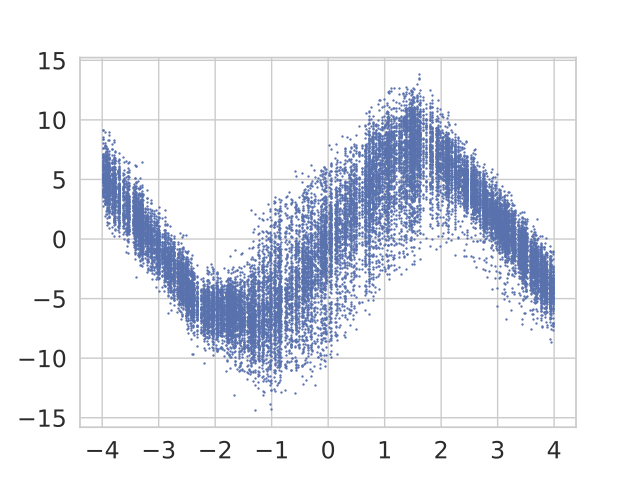} &
    	\hspace{-5mm}
    	\includegraphics[width=3.5cm]{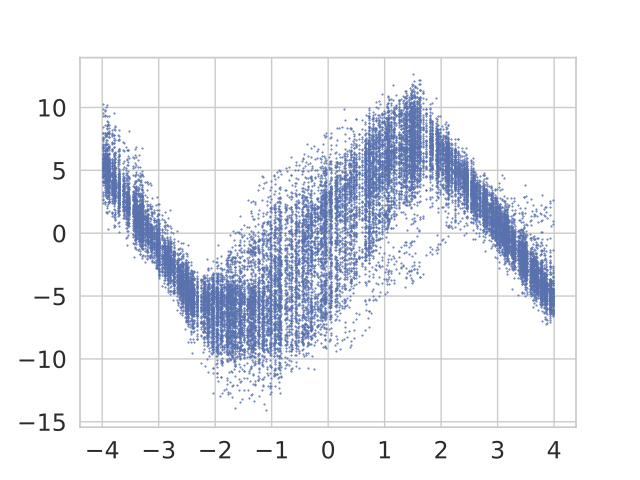}
    	\\
    	\hspace{-5mm}
    	(a) Ground Truth&
    	\hspace{-5mm}
    	(b) VI&
    	\hspace{-5mm}
    	(c) VI ($\alpha=0.5$)&
    	\hspace{-5mm}
    	(d) meta-$\alpha$&
    	\hspace{-5mm}
    	(e) meta-$f$\\
    \end{tabular}
    \caption{Meta-$D$ for BNN regression: visualizing the predictive distributions on sinusoid data. \cz{With our proposed method to meta-learn the divergence (panels (d) and (e)), the learned distribution can accurately capture the uncertainty in different regions while with vanilla VI (panel (b)) or VI with typical $\alpha=0.5$ fails to capture the varying uncertainty in different regions.  }}
    \label{fig:sin-d}
\end{figure*}

The second test considers Bayesian neural network regression. The distribution of ground truth regression function is defined by a sinusoid function with heteroskedastic noise (which is a function of $x$, see Figure \ref{fig:sin-d}(a)):
$y = A\sin(x+b) + A/2\Abs{\cos((x+b)/2)}\epsilon$,
where the amplitude $A\in [5,10]$, the phase $b\in [0,1]$ and $\epsilon\sim\mathcal{N}(0,1)$.
The heteroskedastic noise makes the uncertainty estimate more crucial comparing with the sinusoid function fitting task in prior work \citep{finn2017model,kim2018bayesian}.

For Meta-$D$ ({meta-learning divergences, } Algorithm~\ref{alg:meta-d}), the quantitative results are summarized in Table \ref{tab:sin-d}. We can see that the test log-likelihood (LL) of both meta-$\alpha$ and meta-$f$ are significantly better than VI and VI ($\alpha=0.5$), while the root mean square error (RMSE) are similar for all methods. We visualize the predictive distribution on an example sinusoid function in Figure \ref{fig:sin-d}. All methods fit the mean well which is consistent with the RMSE results. Meta-$\alpha$ and meta-$f$ can reason about the heteroskedastic noise whereas VI and VI ($\alpha=0.5$) used homoskedastic noise to fit the data resulting in bad test LL.

For Meta-$D\&\phi$ (meta-learning divergences and variational parameters, Algorithm~\ref{alg:meta-d-phi}), during meta-testing, we fine-tune the learned $\phi$ with learned divergence on 40 datapoints for 300 epochs. Again meta-$\alpha\&\phi$ and meta-$f\&\phi$ are able to model heteroskedastic predictive distribution while VI$\&\phi$ cannot. The quantitative results are reported in Table \ref{tab:sin-dphi}, and an example of predictive distribution is visualised in Figure \ref{fig:sin-dphi} (see Appendix). Meta-$D\&\phi$ achieves similar results as meta-$D$ with only 40 training data and 300 epochs. Methods without learning initialization for this setup significantly under-perform, indicating that learning model initialization is essential when data is scarce. 

\begin{table}[t]
\begin{minipage}[t]{0.48 \textwidth}
  \centering
  \caption{Meta-$D$ on sin: 10 test tasks and each task has 1000 training data (1000 epochs).}
    \label{tab:sin-d}
    \resizebox{0.8\textwidth}{!}{
    \begin{small}
  \begin{tabular}{ccccc}
  \toprule
    & Test LL & RMSE \\
              \midrule 
  VI &-0.59$\pm$0.01  &0.44$\pm$0.01\\
  VI ($\alpha=0.5$) &-0.57$\pm$0.02  &0.43$\pm$0.01\\
  meta-$\alpha$ &{\bftab -0.39}$\pm$0.04 &0.43$\pm$0.00\\
  meta-$f$ & -0.40$\pm$0.04 &0.42$\pm$0.02 \\
    \bottomrule 
  \end{tabular}
  \end{small}
  }
  \vspace{0.7em}
  \end{minipage}
 \hfill
 \begin{minipage}[t]{0.48 \textwidth}
    \centering
      \caption{Meta-$D\&\phi$ on sin: 10 test tasks and each task has 40 training data (300 epochs).}
  \label{tab:sin-dphi}
  \resizebox{0.8\textwidth}{!}{
  \begin{small}
  \begin{tabular}{ccccc}
  \toprule
    & Test LL & RMSE \\
              \midrule 
    VI &-3.94$\pm$0.18 &0.51$\pm$0.02\\
  VI$\&\phi$ &-0.69$\pm$0.04  &0.44$\pm$0.02\\
  meta-$\alpha\&\phi$ &{\bftab -0.43}$\pm$0.05 &0.42$\pm$0.03\\
  meta-$f\&\phi$ &-0.46$\pm$0.04 &0.43$\pm$0.02 \\
    \bottomrule 
  \end{tabular}
  \end{small}
  }
  \end{minipage}
\end{table}

\subsection{Image Generation with Variational Auto-encoders}\label{sec:exp:mnist}

We also evaluate the image generation task with variational auto-encoders (VAEs).
Specifically, we train VAEs to generate MNIST digits with different divergences. Generating each digit is regarded as a task and we use the first 5 digits (0-4) as the training tasks and the last 5 digits (5-9) as the test tasks.

We report the test marginal log-likelihood for each test digit in Table \ref{tab:mnist-d} and \ref{tab:mnist-dphi}. Overall, these results align with other experiments that the meta-$D$ and meta-$D\&\phi$ are both better than their counterparts. Meta-$D$ and meta-$D\&\phi$ are better than VAE with common divergences on all 5 test tasks, indicating our methods have learned a suitable divergence. 

\begin{table*}[t]
  \caption{Meta-$D$ (meta-learning divergences) on MNIST: marginal log-likelihood on 5 test tasks. Each task has 6000 training data. We train the model for 1000 epochs during meta-testing.}
  \label{tab:mnist-d}
\vspace{-0.7em}
\centering
\resizebox{0.8\textwidth}{!}{
  \begin{tabular}{ccccccc}
  \toprule
    Digit &5 &6 &7 &8 &9 \\
              \midrule
  VI &-133.69$\pm$	0.23 &-121.80$\pm$0.15 &-92.25$\pm$0.40   &-145.14$\pm$0.19  &-119.64$\pm$0.23\\
  VI ($\alpha=0.5$) &-133.24$\pm$0.16 & -121.90$\pm$0.71 & -91.52$\pm$0.72 &-144.90$\pm$0.31 &-119.59$\pm$0.90\\
  meta-$\alpha$ &{\bftab -132.74}$\pm$0.33   &{\bftab -120.67}$\pm$0.36   &{\bftab -90.62} $\pm$0.45   &{ -145.13}$\pm$0.96  &{\bftab -119.42}$\pm$0.36 \\
  meta-$f$ &-133.21$\pm$0.44 &-121.10$\pm$ 0.20 &-91.80$\pm$0.28 &{\bftab -144.85}$\pm$0.31 &{\bftab -119.42}$\pm$0.15\\
    \bottomrule
  \end{tabular}
}
\vspace{0.7em}
  \caption{Meta-$D\&\phi$ (meta-learning divergences and variational parameters) on MNIST: marginal log-likelihood on 5 test tasks. Each task has 100 training data. We train the model for 200 epochs during meta-testing.}
  \label{tab:mnist-dphi}
\vspace{-0.7em}
\resizebox{0.8\textwidth}{!}{
  \begin{tabular}{ccccccc}
  \toprule
    Digit &5 &6 &7 &8 &9 \\
              \midrule 
    VI &-177.92$\pm$0.46 & -182.93$\pm$0.06 & -125.57$\pm$0.41 &-182.63$\pm$0.55&-161.68$\pm$0.27    \\
  VI$\&\phi$ &-174.32$\pm$0.18 &-176.17$\pm$0.26 &-123.20$\pm$0.12   &-177.96$\pm$0.23   &-147.25$\pm$0.32\\

  meta-$\alpha\&\phi$ &{ -163.31}$\pm$0.61   &-163.19$\pm$0.36   &{\bftab -115.52}$\pm$0.16  &{  -173.35}$\pm$0.38  &-142.76$\pm$0.33 \\
  
  meta-$f\&\phi$ &{\bftab -160.16}$\pm$0.16 &{\bftab-154.16}$\pm$0.67 &-122.61$\pm$0.43 &{\bftab -165.83}$\pm$0.48 &{\bftab -138.90}$\pm$0.10\\
    \bottomrule 
  \end{tabular}
}
\end{table*}
\subsection{Recommender System with a Partial Variational Autoencoder}\label{sec:exp:ml}

\begin{figure}[t]
    \hspace{-0.7em}
    \begin{tabular}{cc}
    \includegraphics[width=4.2cm]{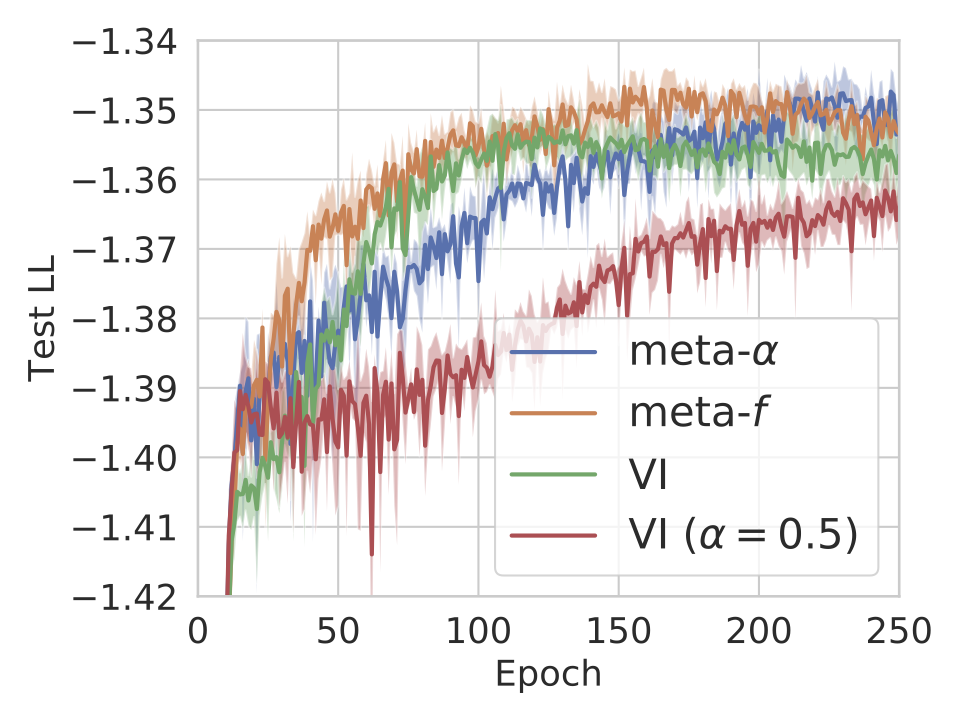}&
    \hspace{-5mm}\includegraphics[width=4.2cm]{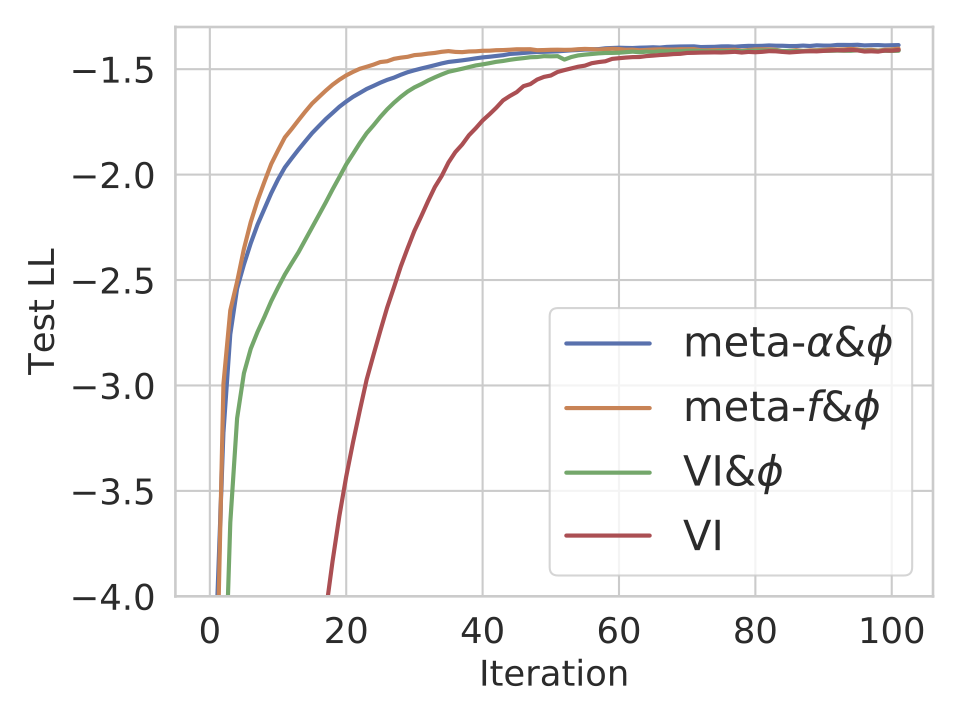}
    \\
    (a) Meta-$D$ &
    (b) Meta-$D\&\phi$ \\
    \end{tabular}
    \caption{Test log-likelihood on MovieLens. \cz{Panel (a) shows the results of meta-learning divergences only (Meta-$D$), and panel (b) shows the results of meta-learning both divergences and variational parameters (Meta-$D\&\phi$). }}
    \label{fig:ml}
\end{figure}

We test our method on recommender systems with a Partial Variational Auto-encoder (p-VAE) \citep{ma2018eddi}. P-VAE is proposed to deal with partially observed data and has been shown to achieve state-of-the-art level performance on user rating prediction in recommender system \citep{ma2018partial}. We consider MovieLens 1M dataset \citep{harper2016movielens} which contains 1,000,206 ratings of 3,952 movies from 6,040 users. We select four age groups as training tasks, and use the remaining three groups as test tasks. During meta-testing, we use 90\%/10\% and 60\%/40\% training-test split for Meta-$D$ and Meta-$D\&\phi$, respectively.
From Figure \ref{fig:ml}(a), we see that when applied to learning p-VAEs, meta-$D$ outperforms standard VI (KL divergence) and VI with $\alpha=0.5$ divergence in terms of test LL, showing that meta-$D$ has learned a suitable divergence that leads to better test performance. Figure \ref{fig:ml}(b) implies that all methods with learned $\phi$ can converge quickly on the new task with only 100 iterations. Both meta-$\alpha\&\phi$ and meta-$f\&\phi$ learn faster than VI$\&\phi$ in meta-test time, indicating that the learned divergence can help fast adaptation.

\section{Related Work}

\paragraph{Variational Inference}
Variational inference (VI) has advanced rapidly in recent years \citep{zhang2018advances}. These advances can be grouped into three categories: (1) introduction of new divergences for VI \citep{bamler2017perturbative,hernandez2016black,li2016renyi}; (2) introduction of more expressive approximate families \citep[e.g.][]{rezende:flow2015,ranganath:hvm2016}; (3) improvement of sampling estimates for model evidence \citep{burda2015importance} and gradient \citep{rainforth2018tighter}; (4) stochastic optimization to scale VI \citep{dehaene2018expectation,hoffman2013stochastic,li2015stochastic}.
Our work is related to the work that improves the variational objective with alternative divergence measures; the difference is that our divergence measure is learnable and can be selected in an automatic fashion for a certain type of tasks.

\paragraph{Meta-Learning/few-shot learning}
Recent work has applied Bayesian modelling techniques to enhance uncertainty estimate for meta-learning/few-shot learning \citep{finn2018probabilistic,grant2018recasting,kim2018bayesian,ravi2018amortized}. They view the framework of MAML \citep{finn2017model} as hierarchical Bayes and conduct Bayesian inference on meta-parameters and/or task-specific parameters. \citet{grant2018recasting} and \citet{kim2018bayesian} applied approximate Bayesian inference to task-specific parameters, while \citet{finn2018probabilistic} kept point estimate for task-specific parameters and conducted variational inference over the meta-parameters instead. \citet{ravi2018amortized} obtained posteriors over both meta and task-specific parameters with variational inference. Our focus is distinct from this line of work in that our research is in the opposite direction: leveraging the idea of meta-learning to advance Bayesian inference. Additionally, our meta-$D\&\phi$ without learning divergence (VI$\&\phi$) can be viewed as a different Bayesian MAML method other than hierarchical Bayes, which directly trains the variational parameters so that it can quickly adapt to new tasks.

\paragraph{Meta-Learning for loss functions} Our meta-learning method is also related to meta-learning a loss function. In reinforcement learning, \citet{houthooft2018evolved} meta-learned the loss function for policy gradients where the parameters of the loss function is updated using evolutionary strategies. \citet{xu2018meta} meta-learned the hyperparameters of the loss functions in TD($\lambda$) and IMPALA. Our work extends the idea of a learnable loss function to Bayesian inference.

\paragraph{Meta-Learning for Bayesian inference algorithms}
A recent attempt to meta-learning stochastic gradient MCMC (SG-MCMC) is presented by \citet{gong2018meta}, which proposed to meta-learn the diffusion and curl matrices of the SG-MCMC's underlying stochastic differential equation.
Also \citet{wang2018meta} applied meta-learning to build efficient and generalizable block-Gibbs sampling proposals.
Our work is distinct from previous work in that we apply meta-learning to improve VI, which is a more scalable inference method than MCMC. To the best of our knowledge, we are the first to study the automatic choice and design of VI algorithms.

\section{Conclusion}
We propose meta-learning divergences of VI which automates the selection of divergence objective in VI via meta-learning. It further allows meta-learning of variational parameter initialization for fast adaptation on new tasks. Within our meta-learning divergences framework, we consider two divergence families, $\alpha$- and $f$-divergence, and design parameterizations of divergences to enable learning via gradient descent. Experimental results on Gaussian mixture approximation, regression with Bayesian neural networks, generative modeling and recommender systems demonstrate the superior performance of meta-learned divergences over standard divergences.

\bibliography{example_paper}

\begin{thebibliography}{44}
\providecommand{\natexlab}[1]{#1}
\providecommand{\url}[1]{\texttt{#1}}
\expandafter\ifx\csname urlstyle\endcsname\relax
  \providecommand{\doi}[1]{doi: #1}\else
  \providecommand{\doi}{doi: \begingroup \urlstyle{rm}\Url}\fi

\bibitem[Amari(2012)]{amari2012differential}
Shun-ichi Amari.
\newblock \emph{Differential-geometrical methods in statistics}, volume~28.
\newblock Springer Science \& Business Media, 2012.

\bibitem[Amit and Meir(2018)]{amit2018meta}
Ron Amit and Ron Meir.
\newblock Meta-learning by adjusting priors based on extended {PAC}-{B}ayes
  theory.
\newblock In \emph{International Conference on Machine Learning}, pages
  205--214. PMLR, 2018.

\bibitem[Bamler et~al.(2017)Bamler, Zhang, Opper, and
  Mandt]{bamler2017perturbative}
Robert Bamler, Cheng Zhang, Manfred Opper, and Stephan Mandt.
\newblock Perturbative black box variational inference.
\newblock In \emph{Advances in Neural Information Processing Systems}, pages
  5079--5088, 2017.

\bibitem[Bishop(2006)]{bishop2006pattern}
Christopher~M Bishop.
\newblock \emph{Pattern recognition and machine learning}.
\newblock Springer Science+ Business Media, 2006.

\bibitem[Blei et~al.(2017)Blei, Kucukelbir, and McAuliffe]{blei2017variational}
David~M Blei, Alp Kucukelbir, and Jon~D McAuliffe.
\newblock Variational inference: A review for statisticians.
\newblock \emph{Journal of the American Statistical Association}, 112\penalty0
  (518):\penalty0 859--877, 2017.

\bibitem[Blundell et~al.(2015)Blundell, Cornebise, Kavukcuoglu, and
  Wierstra]{blundell2015weight}
Charles Blundell, Julien Cornebise, Koray Kavukcuoglu, and Daan Wierstra.
\newblock Weight uncertainty in neural networks.
\newblock \emph{International Conference on Machine Learning}, 2015.

\bibitem[Burda et~al.(2015)Burda, Grosse, and
  Salakhutdinov]{burda2015importance}
Yuri Burda, Roger Grosse, and Ruslan Salakhutdinov.
\newblock Importance weighted autoencoders.
\newblock \emph{arXiv preprint arXiv:1509.00519}, 2015.

\bibitem[Cichocki and Amari(2010)]{cichocki2010families}
Andrzej Cichocki and Shun-ichi Amari.
\newblock Families of alpha-beta-and gamma-divergences: Flexible and robust
  measures of similarities.
\newblock \emph{Entropy}, 12\penalty0 (6):\penalty0 1532--1568, 2010.

\bibitem[Csisz{\'a}r et~al.(2004)Csisz{\'a}r, Shields,
  et~al.]{csiszar2004information}
Imre Csisz{\'a}r, Paul~C Shields, et~al.
\newblock Information theory and statistics: A tutorial.
\newblock \emph{Foundations and Trends{\textregistered} in Communications and
  Information Theory}, 1\penalty0 (4):\penalty0 417--528, 2004.

\bibitem[Dehaene and Barthelm{\'e}(2018)]{dehaene2018expectation}
Guillaume Dehaene and Simon Barthelm{\'e}.
\newblock Expectation propagation in the large data limit.
\newblock \emph{Journal of the Royal Statistical Society: Series B (Statistical
  Methodology)}, 80\penalty0 (1):\penalty0 199--217, 2018.

\bibitem[Depeweg et~al.(2017)Depeweg, Hern{\'a}ndez-Lobato, Doshi-Velez, and
  Udluft]{depeweg2016learning}
Stefan Depeweg, Jos{\'e}~Miguel Hern{\'a}ndez-Lobato, Finale Doshi-Velez, and
  Steffen Udluft.
\newblock Learning and policy search in stochastic dynamical systems with
  {B}ayesian neural networks.
\newblock \emph{International Conference on Learning Representations}, 2017.

\bibitem[Finn et~al.(2017)Finn, Abbeel, and Levine]{finn2017model}
Chelsea Finn, Pieter Abbeel, and Sergey Levine.
\newblock Model-agnostic meta-learning for fast adaptation of deep networks.
\newblock In \emph{Proceedings of the 34th International Conference on Machine
  Learning-Volume 70}, pages 1126--1135. JMLR. org, 2017.

\bibitem[Finn et~al.(2018)Finn, Xu, and Levine]{finn2018probabilistic}
Chelsea Finn, Kelvin Xu, and Sergey Levine.
\newblock Probabilistic model-agnostic meta-learning.
\newblock In \emph{Advances in Neural Information Processing Systems}, pages
  9516--9527, 2018.

\bibitem[Flennerhag et~al.(2019)Flennerhag, Moreno, Lawrence, and
  Damianou]{flennerhag2018transferring}
Sebastian Flennerhag, Pablo~G Moreno, Neil~D Lawrence, and Andreas Damianou.
\newblock Transferring knowledge across learning processes.
\newblock \emph{International Conference on Learning Representations}, 2019.

\bibitem[Flennerhag et~al.(2020)Flennerhag, Rusu, Pascanu, Yin, and
  Hadsell]{flennerhag2019meta}
Sebastian Flennerhag, Andrei~A Rusu, Razvan Pascanu, Hujun Yin, and Raia
  Hadsell.
\newblock Meta-learning with warped gradient descent.
\newblock \emph{International Conference on Learning Representations}, 2020.

\bibitem[Gilardoni(2010)]{gilardoni2010pinsker}
Gustavo~L Gilardoni.
\newblock On {P}insker's and {V}ajda's type inequalities for {C}sisz{\'a}r's $
  f $-divergences.
\newblock \emph{IEEE Transactions on Information Theory}, 56\penalty0
  (11):\penalty0 5377--5386, 2010.

\bibitem[Gong et~al.(2019)Gong, Li, and Hern{\'a}ndez-Lobato]{gong2018meta}
Wenbo Gong, Yingzhen Li, and Jos{\'e}~Miguel Hern{\'a}ndez-Lobato.
\newblock Meta-learning for stochastic gradient {MCMC}.
\newblock \emph{International Conference on Learning Representations}, 2019.

\bibitem[Grant et~al.(2018)Grant, Finn, Levine, Darrell, and
  Griffiths]{grant2018recasting}
Erin Grant, Chelsea Finn, Sergey Levine, Trevor Darrell, and Thomas Griffiths.
\newblock Recasting gradient-based meta-learning as hierarchical {B}ayes.
\newblock \emph{International Conference on Learning Representations}, 2018.

\bibitem[Harper and Konstan(2016)]{harper2016movielens}
F~Maxwell Harper and Joseph~A Konstan.
\newblock The movielens datasets: History and context.
\newblock \emph{Acm transactions on interactive intelligent systems (tiis)},
  5\penalty0 (4):\penalty0 19, 2016.

\bibitem[Hern{\'a}ndez-Lobato et~al.(2016)Hern{\'a}ndez-Lobato, Li, Rowland,
  Hern{\'a}ndez-Lobato, Bui, and Turner]{hernandez2016black}
Jos{\'e}~Miguel Hern{\'a}ndez-Lobato, Yingzhen Li, Mark Rowland, Daniel
  Hern{\'a}ndez-Lobato, Thang Bui, and Richard Turner.
\newblock Black-box $\alpha$-divergence minimization.
\newblock \emph{International Conference on Machine Learning,}, 2016.

\bibitem[Hoffman et~al.(2013)Hoffman, Blei, Wang, and
  Paisley]{hoffman2013stochastic}
Matthew~D Hoffman, David~M Blei, Chong Wang, and John Paisley.
\newblock Stochastic variational inference.
\newblock \emph{The Journal of Machine Learning Research}, 14\penalty0
  (1):\penalty0 1303--1347, 2013.

\bibitem[Houthooft et~al.(2018)Houthooft, Chen, Isola, Stadie, Wolski, Ho, and
  Abbeel]{houthooft2018evolved}
Rein Houthooft, Yuhua Chen, Phillip Isola, Bradly Stadie, Filip Wolski,
  OpenAI~Jonathan Ho, and Pieter Abbeel.
\newblock Evolved policy gradients.
\newblock In \emph{Advances in Neural Information Processing Systems}, pages
  5400--5409, 2018.

\bibitem[Jordan et~al.(1999)Jordan, Ghahramani, Jaakkola, and
  Saul]{jordan1999introduction}
Michael~I Jordan, Zoubin Ghahramani, Tommi~S Jaakkola, and Lawrence~K Saul.
\newblock An introduction to variational methods for graphical models.
\newblock \emph{Machine learning}, 37\penalty0 (2):\penalty0 183--233, 1999.

\bibitem[Kim et~al.(2018)Kim, Yoon, Dia, Kim, Bengio, and Ahn]{kim2018bayesian}
Taesup Kim, Jaesik Yoon, Ousmane Dia, Sungwoong Kim, Yoshua Bengio, and Sungjin
  Ahn.
\newblock Bayesian model-agnostic meta-learning.
\newblock \emph{Advances in Neural Information Processing Systems}, 2018.

\bibitem[Kingma and Welling(2014)]{kingma2013auto}
Diederik~P Kingma and Max Welling.
\newblock Auto-encoding variational {B}ayes.
\newblock \emph{International Conference on Learning Representations4}, 2014.

\bibitem[Li and Turner(2016)]{li2016renyi}
Yingzhen Li and Richard~E Turner.
\newblock R{\'e}nyi divergence variational inference.
\newblock In \emph{Advances in Neural Information Processing Systems}, pages
  1073--1081, 2016.

\bibitem[Li et~al.(2015)Li, Hern{\'a}ndez-Lobato, and Turner]{li2015stochastic}
Yingzhen Li, Jos{\'e}~Miguel Hern{\'a}ndez-Lobato, and Richard~E Turner.
\newblock Stochastic expectation propagation.
\newblock In \emph{Advances in neural information processing systems}, pages
  2323--2331, 2015.

\bibitem[Ma et~al.(2018)Ma, Gong, Hern{\'a}ndez-Lobato, Koenigstein, Nowozin,
  and Zhang]{ma2018partial}
Chao Ma, Wenbo Gong, Jos{\'e}~Miguel Hern{\'a}ndez-Lobato, Noam Koenigstein,
  Sebastian Nowozin, and Cheng Zhang.
\newblock Partial {VAE} for hybrid recommender system.
\newblock In \emph{NIPS Workshop on Bayesian Deep Learning}, 2018.

\bibitem[Ma et~al.(2019)Ma, Tschiatschek, Palla, Lobato, Nowozin, and
  Zhang]{ma2018eddi}
Chao Ma, Sebastian Tschiatschek, Konstantina Palla, Jose Miguel~Hernandez
  Lobato, Sebastian Nowozin, and Cheng Zhang.
\newblock Eddi: Efficient dynamic discovery of high-value information with
  partial {VAE}.
\newblock \emph{International Conference on Machine Learning}, 2019.

\bibitem[Minka(2001)]{minka2001expectation}
Thomas~P Minka.
\newblock Expectation propagation for approximate {B}ayesian inference.
\newblock In \emph{Proceedings of the Seventeenth conference on Uncertainty in
  artificial intelligence}, pages 362--369. Morgan Kaufmann Publishers Inc.,
  2001.

\bibitem[Minka et~al.(2005)]{minka2005divergence}
Tom Minka et~al.
\newblock Divergence measures and message passing.
\newblock Technical report, Technical report, Microsoft Research, 2005.

\bibitem[Rainforth et~al.(2018)Rainforth, Kosiorek, Le, Maddison, Igl, Wood,
  and Teh]{rainforth2018tighter}
Tom Rainforth, Adam~R Kosiorek, Tuan~Anh Le, Chris~J Maddison, Maximilian Igl,
  Frank Wood, and Yee~Whye Teh.
\newblock Tighter variational bounds are not necessarily better.
\newblock \emph{International Conference on Machine Learning}, 2018.

\bibitem[Rajeswaran et~al.(2019)Rajeswaran, Finn, Kakade, and
  Levine]{rajeswaran2019meta}
Aravind Rajeswaran, Chelsea Finn, Sham~M Kakade, and Sergey Levine.
\newblock Meta-learning with implicit gradients.
\newblock In \emph{Advances in Neural Information Processing Systems}, pages
  113--124, 2019.

\bibitem[Ranganath et~al.(2016)Ranganath, Tran, and Blei]{ranganath:hvm2016}
Rajesh Ranganath, Dustin Tran, and David Blei.
\newblock Hierarchical variational models.
\newblock In \emph{Proceedings of the 33rd International Conference on Machine
  Learning}, pages 324--333, 2016.

\bibitem[Ravi and Beatson(2019)]{ravi2018amortized}
Sachin Ravi and Alex Beatson.
\newblock Amortized {B}ayesian meta-learning.
\newblock \emph{International Conference on Learning Representations}, 2019.

\bibitem[R{\'e}nyi et~al.(1961)]{renyi1961measures}
Alfr{\'e}d R{\'e}nyi et~al.
\newblock On measures of entropy and information.
\newblock In \emph{Proceedings of the Fourth Berkeley Symposium on Mathematical
  Statistics and Probability, Volume 1: Contributions to the Theory of
  Statistics}. The Regents of the University of California, 1961.

\bibitem[Rezende and Mohamed(2015)]{rezende:flow2015}
Danilo Rezende and Shakir Mohamed.
\newblock Variational inference with normalizing flows.
\newblock In \emph{Proceedings of the 32nd International Conference on Machine
  Learning}, pages 1530--1538, 2015.

\bibitem[Salimans et~al.(2013)Salimans, Knowles, et~al.]{salimans2013fixed}
Tim Salimans, David~A Knowles, et~al.
\newblock Fixed-form variational posterior approximation through stochastic
  linear regression.
\newblock \emph{Bayesian Analysis}, 8\penalty0 (4):\penalty0 837--882, 2013.

\bibitem[Snoek et~al.(2012)Snoek, Larochelle, and Adams]{snoek2012practical}
Jasper Snoek, Hugo Larochelle, and Ryan~P Adams.
\newblock Practical {B}ayesian optimization of machine learning algorithms.
\newblock In \emph{Advances in neural information processing systems}, pages
  2951--2959, 2012.

\bibitem[Tsallis(1988)]{tsallis1988possible}
Constantino Tsallis.
\newblock Possible generalization of {B}oltzmann-{G}ibbs statistics.
\newblock \emph{Journal of statistical physics}, 52\penalty0 (1-2):\penalty0
  479--487, 1988.

\bibitem[Wang et~al.(2018{\natexlab{a}})Wang, Liu, and
  Liu]{wang2018variational}
Dilin Wang, Hao Liu, and Qiang Liu.
\newblock Variational inference with tail-adaptive f-divergence.
\newblock In \emph{Advances in Neural Information Processing Systems}, pages
  5737--5747, 2018{\natexlab{a}}.

\bibitem[Wang et~al.(2018{\natexlab{b}})Wang, Wu, Moore, and
  Russell]{wang2018meta}
Tongzhou Wang, Yi~Wu, Dave Moore, and Stuart~J Russell.
\newblock Meta-learning {MCMC} proposals.
\newblock In \emph{Advances in Neural Information Processing Systems}, pages
  4146--4156, 2018{\natexlab{b}}.

\bibitem[Xu et~al.(2018)Xu, van Hasselt, and Silver]{xu2018meta}
Zhongwen Xu, Hado~P van Hasselt, and David Silver.
\newblock Meta-gradient reinforcement learning.
\newblock In \emph{Advances in neural information processing systems}, pages
  2396--2407, 2018.

\bibitem[Zhang et~al.(2018)Zhang, Butepage, Kjellstrom, and
  Mandt]{zhang2018advances}
Cheng Zhang, Judith Butepage, Hedvig Kjellstrom, and Stephan Mandt.
\newblock Advances in variational inference.
\newblock \emph{IEEE transactions on pattern analysis and machine
  intelligence}, 2018.

\end{thebibliography}
\bibliographystyle{plainnat}

\clearpage
\appendix
\newpage
\onecolumn

\section{Computing Equation (\ref{eq:f-grad}) in Practice}\label{sec:compute-f-grad}
With dataset $\mathcal{D}$, the density ratio in $f$-divergence becomes $\frac{p(\theta | D)}{q_{\phi}(\theta)} = \frac{p(D | \theta)p(\theta)}{q_{\phi}(\theta)p(D)}$. We estimate $p(D)$ through importance sampling and MC approximation: $p(D) = E_{\theta\sim p(\theta)} [p(D | \theta)] = E_{\theta\sim q_{\phi}(\theta)} [\frac{p(D | \theta)p(\theta)}{q_{\phi}(\theta)}] \approx \frac{1}{K}\sum_{k=1}^K \frac{p(D | \theta_k)p(\theta_k)}{q_{\phi}(\theta_k)}$ where $\theta_k \sim q_{\phi}(\theta)$. After doing this, the density ratio becomes $\frac{p(\theta_k | D)}{q_{\phi}(\theta_k)} = \frac{p(D | \theta_k)p(\theta_k)}{q_{\phi}(\theta_k)}\bigg/ \frac{1}{K}\sum_{k=1}^K \frac{p(D | \theta_k)p(\theta_k)}{q_{\phi}(\theta_k)}$ which can be regarded as a self-normalized estimator, similar to the normalization importance weight in \citet{li2016renyi}. A self-normalized estimator generally helps stabilize the training especially at the beginning. We use Eq.(\ref{eq:f-grad}) with this estimator and stochastic approximation of gradients for all experiments except for the mixture of Gaussians task where we can directly compute $p(\theta)/q_{\phi}(\theta)$.

\section{Effect of Hyperparameter $B$}\label{app:hyperparameter}
Similar to other MAML-based algorithms~\citep{finn2017model,finn2018probabilistic,kim2018bayesian}, the cost of our method increases as hyperparameter $B$ increases and the value of $B$ could potentially affect the results. As in prior work, we treat $B$ as a hyperparameter and tune it for each task. Empirically, we found setting $B =1$ is enough for most tasks we considered in the experimental section. For example, we also tried $B=2, 5$ for meta-$\alpha$ in Section \ref{sec:exp:sin} and found that they gave similar values of learned $\alpha$ as $B=1$ ($\alpha=0.10, 0.13, 0.16$ for $B=1, 2, 5$ respectively). Setting $B$ larger will be costly and even cause gradients to be problematic due to requiring higher-order derivatives. We may combine our methods with recent techniques in meta-learning~\citep{flennerhag2018transferring,flennerhag2019meta,rajeswaran2019meta} to allow large $B$, which is an interesting future work.  

\section{Additional Experimental Results and Setting Details}\label{app:exp}

\subsection{Task Distribution $p(\mathcal{T})$}
When the number of training tasks is finite (which is often the case in practice) such as image generation with MNIST (Section~\ref{sec:exp:mnist}) and recommender system with MovieLens (Section~\ref{sec:exp:ml}), the task distribution $p(\mathcal{T})$ is defined as a uniform distribution over all training tasks for both meta-$D$ (Algorithm~\ref{alg:meta-d}) and meta-$D\&\phi$ (Algorithm~\ref{alg:meta-d-phi}). When the number of training tasks is infinite such as Gaussian mixture approximation (Section~\ref{sec:exp:gmm}) and sinusoid regression (Section~\ref{sec:exp:sin}), we use a uniform distribution over all training tasks as $p(\mathcal{T})$ for meta-$D\&\phi$ but a uniform distribution over a subsampled set of training tasks as $p(\mathcal{T})$ for meta-$D$ (the set size is 10 and 20 for Gaussian mixture approximation and sinusoid regression respectively). This is to avoid storing too many models since meta-$D$ allows each task $T_i$ has its own model $\phi_i$.

\subsection{Parameterization of $f$-Divergence in Practice}
Based on the Proposition \ref{prop:f-gradient} and \ref{prop:2}, we can parameterize $f$-divergence by parameterizing $g(t) = t^2f''(t) = \exp(h_{\eta}(t))$ where $h_{\eta}$ is a neural network with parameter $\eta$. However, this way of parameterization makes it hard to learn the divergences whose $g(t)$ is very small when $t$ is small (because $h(t)$ has to output negative numbers with large absolute values), such as Renyi divergence with $\alpha\approx 0$. These kinds of divergences behave like approximating the expectation in Eq.(\ref{eq:f-grad}) only with $\theta$ whose $p(\theta)/q_{\phi}(\theta)$ is large, which is important for modeling bimodal and heteroscedastic distributions~\citep{depeweg2016learning}.  

To alleviate this issue, we can instead parameterize $f''(t) = \exp(h_{\eta}(t))$, then $g(t)$ in Eq.(\ref{eq:f-grad}) becomes $t^2\exp(h_{\eta}(t))$. It is easy to see that this parameterization solves the above issue due to $t^2$ which becomes small when $t$ is small. However, it is hard to learn the divergences that put similar weights to MC samples (e.g. standard KL$(q\|p)$, which gives the equal weights). These two ways of parameterization are statistically equivalent but have different inductive bias. Parameterizing $f''$ tends to learn a divergence that puts different weights to MC samples according to $p(\theta)/q_{\phi}(\theta)$ (due to $t^2$). On the other hand, parameterizing $g(t)$ tends to give relatively similar weights to MC samples. In the experiments, we parameterize $g(t)$ when the learned $\alpha$ from meta-$\alpha$ is close to 1 (Section \ref{sec:exp:gmm} and \ref{sec:exp:ml}) and parameterize $f''(t)$ when the learned $\alpha$ is close to 0 (Section \ref{sec:exp:sin} and \ref{sec:exp:mnist}). We found this strategy works well in practice.

\subsection{Model Architecture for $f$-divergence}
On all experiments, we parameterize $h_{\eta}(t)$ in $f$-divergence by a neural network with 2 hidden layers with 100 hidden units and RELU nonlinearilities. In practice, we find that pretraining $h_{\eta}(t)$ to be the standard KL divergence can stabilize the training at the beginning. We initialize $h_{\eta}(t)$ in this way for all experiments.

\subsection{Approximate Mixture of Gaussians}\label{app:mog}
\begin{figure*}[t]
    \centering
    \begin{tabular}{cccc}
    \includegraphics[width=3.8cm]{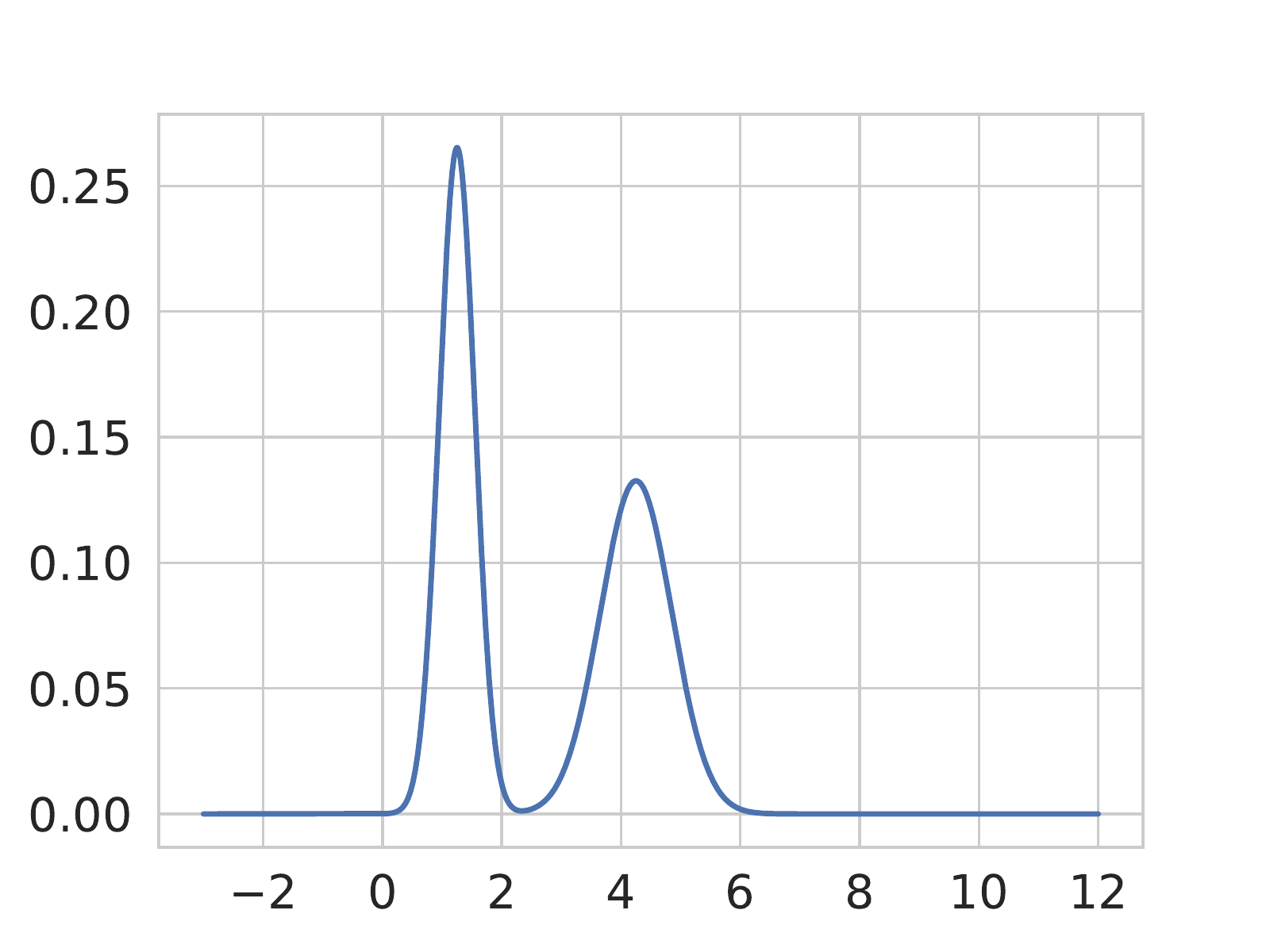}&
    	\includegraphics[width=3.8cm]{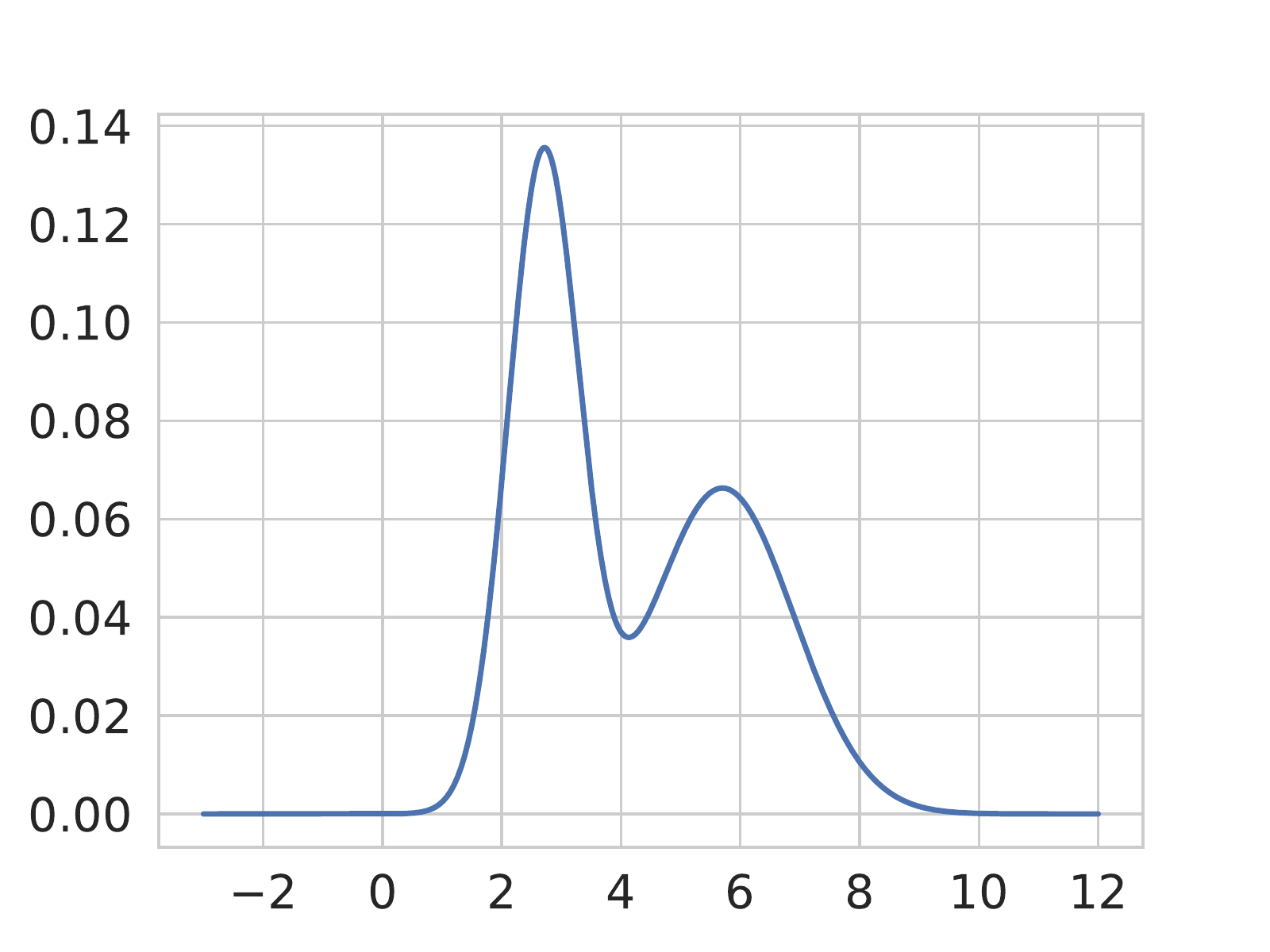} &
    	\includegraphics[width=3.8cm]{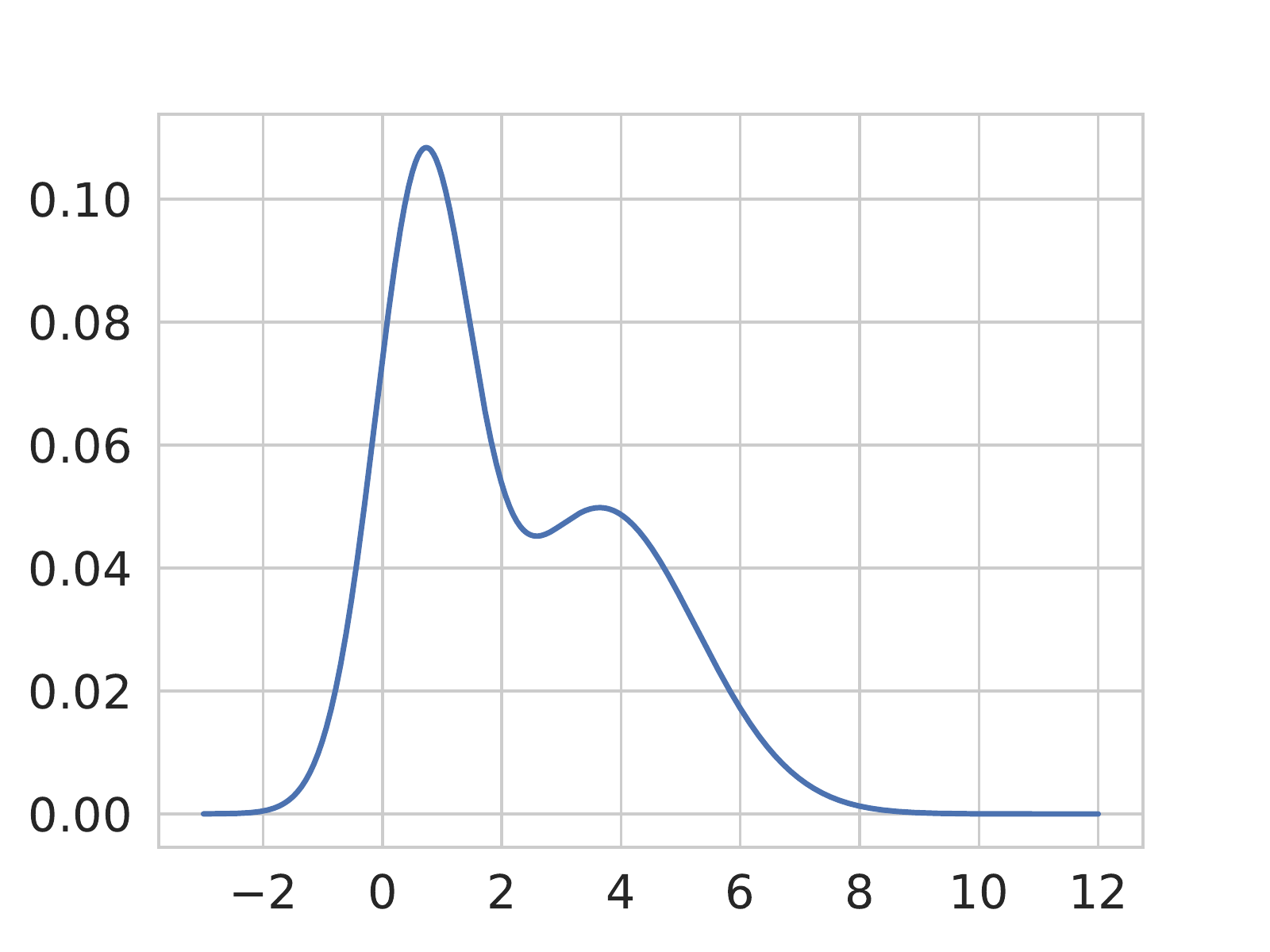} 
    	\\
    \end{tabular}
    \caption{Three examples of mixture of Gaussians. Each task includes approximating a mixture of Gaussians by a Gaussian distribution.}
    \label{fig:gmm_example}
\end{figure*}

In this experiment, each task is to approximate a mixture of Gaussians by a Gaussian distribution. We give examples of the mixture of Gaussians in Figure~\ref{fig:gmm_example}. The expectation in Eq.(\ref{eq:vr-bound}) and (\ref{eq:f-grad}) is computed by MC approximation with 1000 particles. Note that $p(\theta)$ is computable, since we know the parameters of $p$. 

\subsubsection{TV Distance}

TV is a common distance measure for probability distributions. It is defined as
\begin{align*}
\operatorname{TV}(p, q) &= \sup_x |p(x) - q(x)| = \frac{1}{2} \int |p(x) - q(x)| \; dx.
\end{align*}
For $\alpha \in (0, 1]$, TV is related to $\alpha$-divergence by
$
\alpha \operatorname{TV}^2 \le 2 \cdot D_{\alpha}(p\|q)
$ \citep{gilardoni2010pinsker}.

Note that although TV belongs to a more general $f$-divergence family by setting $f_{\text{tv}}(t)= \Abs{t-1}/2$, it does not belong to the $f$-divergence we defined in the paper. Since $f_{\text{TV}}(t)$ is not twice-differentiable. Therefore, we can use TV as an example to test the performance of our methods when meta-loss is beyond $\alpha$- and $f$-divergence.  
\subsubsection{Bayesian Optimization}
We used a standard setup of BO, following \cite{snoek2012practical}. To ensure fair comparisons, we implemented BO through a public and stable library \footnote{\url{https://github.com/fmfn/BayesianOptimization}.}. We set the search region for BO to be $\alpha \in [0, 3]$. The acquisition function is the upper confidence bound with kappa 0.1. We used the same data for training meta-$D$ for BO. Specifically, the objective function that BO minimizes is the meta-loss ($D_{0.5}$ or TV). Every time BO selects an $\alpha$, we train 10 models with that $\alpha$-divergence on the training sets of 10 training tasks respectively and get the mean of log-likelihood on the test sets of the 10 training tasks. Each time the model is trained for 2000 iterations. It is possible to choose the best $\alpha$ for each test task by BO, i.e. every time we have a new test task we run BO to select an $\alpha$ for this task. However, by doing this, we are not able to extract any common knowledge from the previous tasks and running BO for each task could be very expensive. We did not include this baseline because it does not satisfy the meta-learning setting and the cost will be much higher than our methods. 
\subsubsection{Analytical Expression of $h_f(t)$}

When $f$-divergence is $D_{0.5}$, the $f$ function is $f(t)=\frac{t^{0.5}}{-0.5^2}$. Then we can write out the corresponding $h_f(t)$ as
\[
h_f(t) =\log g_f(t) = \log f''(t) + 2\log t =  0.5 \log t.
\]
Because the definition of $f$-divergence is invariant to constant scaling of the function $f$, i.e.~$f$ and $a f$ define the same divergence for $\forall a>0$, we consider the corresponding $h_f(t)$ for $af$ which is
\[
h_f(t) =  0.5 \log t + \log a.
\]
In Figure 2, we compare the learned $h_{\eta}(t)$ and the ground truth $h_f(t)$. We found that the learned $h_{\eta}(t)$ is very close to $0.5 \log t + 1.25$, which means that our method has learned the optimal divergence $D_{0.5}$. We conjecture that the constant $a$ for the learned $f$ is related to the learning rate. In fact, this gives $f$-divergence the ability to automatically adjust the learning rate through the scaling constant. For example, the constant $a$ is $\exp(1.25)>1$ which may suggest that the learning rate $\beta$ is a bit small. Note that $\alpha$-divergence does not have this ability. We plot $f$ for $t\in [0, 3]$ in Figure~\ref{fig:gmm-dphi} since we find most $t$ lie in this range. 

\subsubsection{Additional Experimental Results}
For meta-$D$, we report in Table \ref{tab:value-metaloss} the meta-losses on 10 test tasks, which are obtained by executing the learned divergence minimization algorithm for 2000 iterations. The error bar is large due to the large variance among different tasks, so we report the ranking in Table \ref{tab:rank-d}, similar to \citet{ma2018eddi}, to clearly show the adavantages of meta-$D$ over BO. Similarly, we also report the meta-losses for meta-$D\&\phi$ over 10 tasks in Table~\ref{tab:meta-loss-dphi}.

\begin{table}[h]
\caption{Meta-$D$ on MoG: value of meta-loss over 10 test tasks.}
\label{tab:value-metaloss}
\begin{center}
\begin{tabular}{lcccc}
\toprule
Methods & $\alpha=0.5$ &TV\\
\midrule
ground truth &0.0811$\pm$0.0277   & -\\
  meta-$\alpha$ &0.0811$\pm$0.0277 & 0.2143$\pm$0.0936\\
  meta-$f$ &{\bftab 0.0795}$\pm$0.0301 &{\bftab 0.2020}$\pm$0.1024 \\
  BO (8 iters) &0.0833$\pm$0.0289 &0.2203$\pm$0.0898 \\
  BO (16 iters) &0.0811$\pm$0.0277 & 0.2143$\pm$0.0936\\
\bottomrule
\end{tabular}
\end{center}
\end{table}

\addtolength{\tabcolsep}{-3pt} 
\begin{table*}[h]
  \centering
  \caption{Meta-$D$\&$\phi$ on MoG: value of meta-loss over 10 test tasks.}
  \label{tab:meta-loss-dphi}
  \begin{tabular}{ccc|cc}
  \toprule
    Methods\textbackslash Meta-loss & $\alpha=0.5$ (20 iters) &TV (20 iters)&$\alpha=0.5$ (100 iters)&TV (100 iters)\\
    \midrule
  VI\&$\phi$ &0.1237$\pm$0.0539 &0.2572$\pm$0.1137 &0.0905$\pm$0.0332 &0.2321$\pm$0.0961\\
  meta-$\alpha\&\phi$ &0.1207$\pm$0.0500 &0.2462$\pm$0.1043 &0.0879$\pm$0.0305 &{\bftab 0.2263}$\pm$0.0936 \\
  meta-$f\&\phi$ &{\bftab 0.0793}$\pm$0.0237 &{\bftab 0.2344}$\pm$0.0955 &{\bftab 0.0784}$\pm$0.0332 &0.2301$\pm$0.0949\\
    \bottomrule
  \end{tabular}
\end{table*}
\addtolength{\tabcolsep}{3pt} 
  
\subsection{Regression Tasks with Bayesian Neural Networks}
\label{app:BNNsetting}
\begin{figure*}[h]
    \centering
    \begin{tabular}{cccc}
    \includegraphics[width=3.8cm]{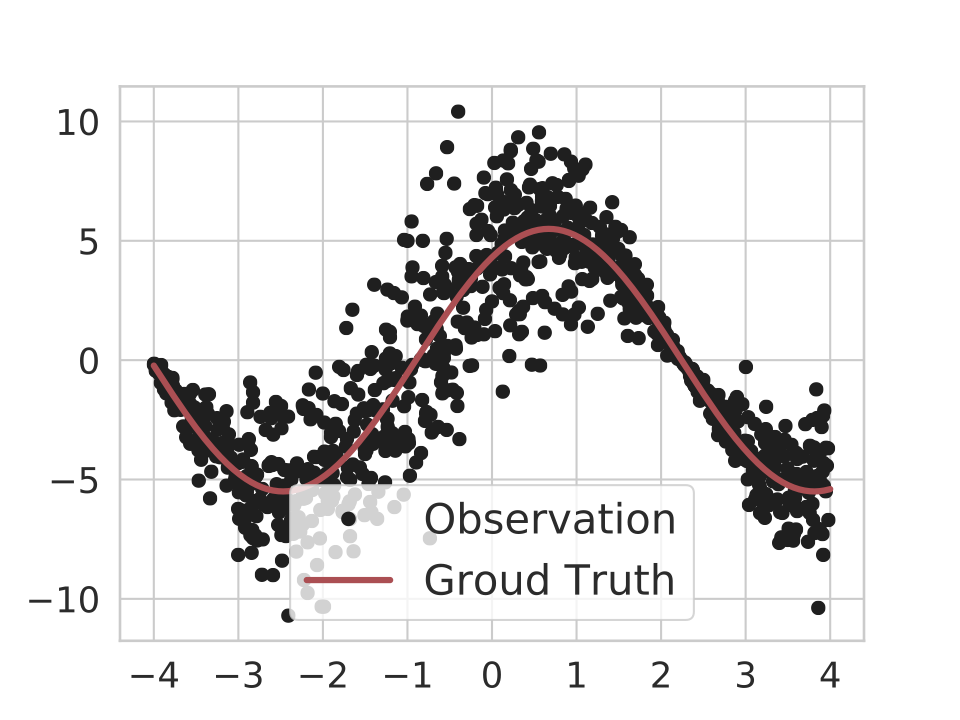}&
    	\includegraphics[width=3.8cm]{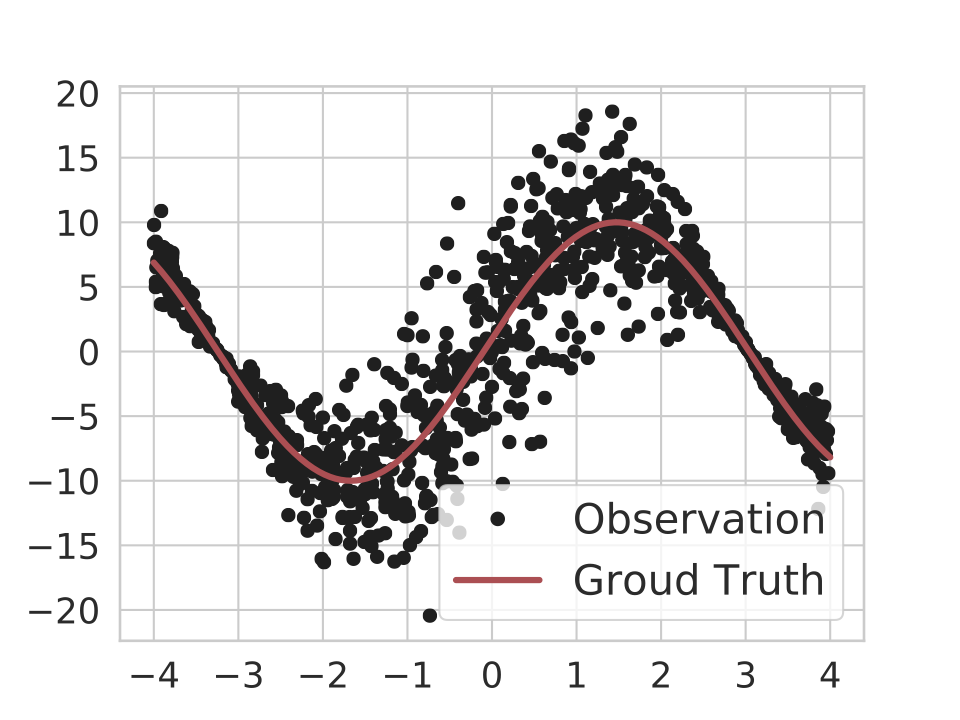} &
    	\includegraphics[width=3.8cm]{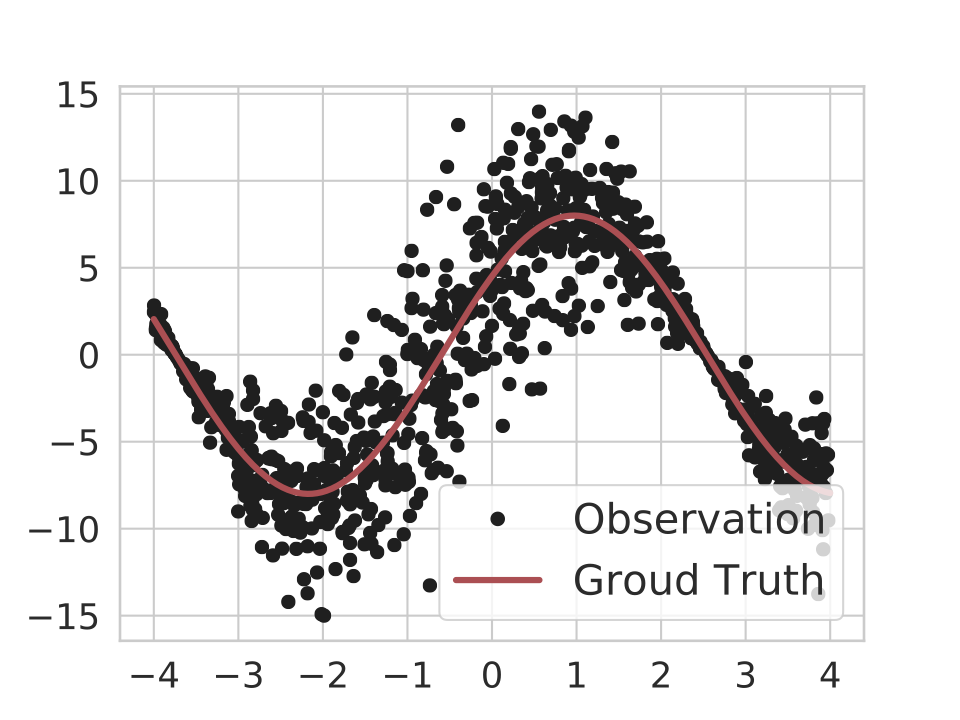} 
    	\\
    \end{tabular}
    \caption{Three examples of sinusoid waves. Each task includes a regression on a sinusoid wave.}
    \label{fig:sin_example}
\end{figure*}

Each task includes a regression problem on a sinusoid wave; see Figure~\ref{fig:sin_example} for examples of the sinusoid waves.
The BNN model is a two-layer neural network with hidden layer size 20 and RELU nonlinearities. 
For meta-learning divergence only, the training set size is 1000 and is obtained by sampling $x\in [-4,4]$ uniformly. We use $K=100$ and batch size 40 of which 20 data points are for updating $\phi_i$ and 20 points are for updating $\eta$. We train meta-$D$ for 1000 epochs. To evaluate the performance, we train the model with the learned divergence and VI respectively on new tasks for 1000 epochs. For learning both the divergence objective and initial variational parameters, we sample 20 tasks each time where each task has 40 data points. We use 20 points for updating $\phi_i$ and the other 20 points for updating divergence $\eta$ and the shared initialization $\phi$. $B=4$ for meta-$\alpha\&\phi$. To evaluate, we start with the learned initialization and train the variational parameters with the learned divergence for 300 epochs.
\subsubsection{Additional Experimental Results}
\begin{figure*}[h]
    \centering
    \begin{tabular}{cccc}
    \includegraphics[width=3.2cm]{figs/sin/sin_alpha_truth.png}  &
    	\includegraphics[width=3.2cm]{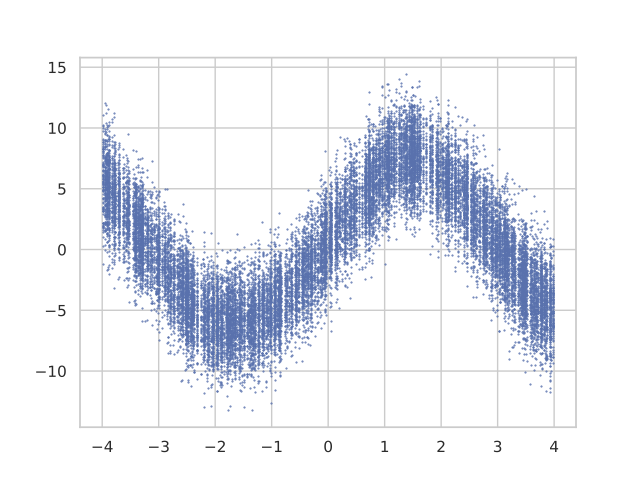} &
    	\includegraphics[width=3.2cm]{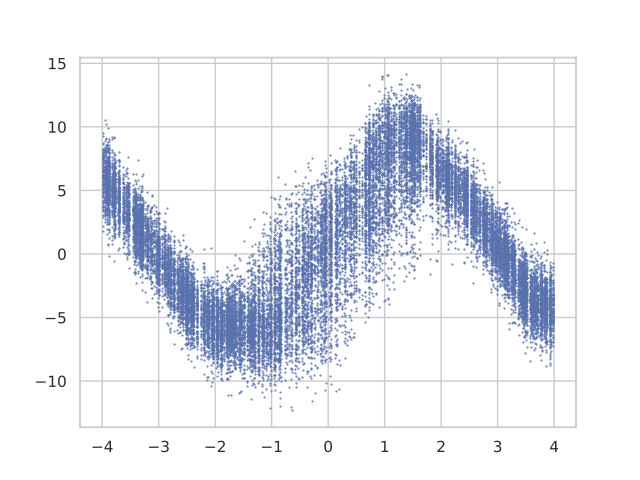} &
    	\includegraphics[width=3.2cm]{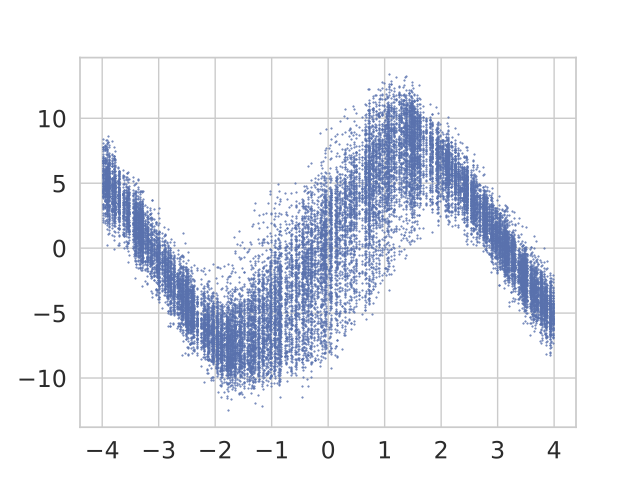}
    	\\
    	(a) Ground Truth&
    	(b) VI\&$\phi$&
    	(c) meta-$\alpha$\&$\phi$&
    	(d) meta-$f\&\phi$\\
    \end{tabular}
    \caption{Meta-$D\&\phi$ on sin: the predictive distribution on a sinusoid wave.}
    \label{fig:sin-dphi}
\end{figure*}

We provide the learned value of $\alpha$ from meta-$\alpha$ and meta-$\alpha\&\phi$ in Table \ref{fig:learned-alpha-sin}. The predictive distributions on an example test task are given in Figure \ref{fig:sin-dphi}. Similar to the results of meta-$D$, meta-$D\&\phi$ is also able to model heteroskedastic noise while VI\&$\phi$ cannot.

\begin{table}[H]
\begin{minipage}[t]{0.48 \textwidth}
  \centering
    \caption{Learned value of $\alpha$ of meta-$\alpha$ and meta-$\alpha\&\phi$ on sinusoid regression.}
    \label{fig:learned-alpha-sin}
\begin{center}
  \begin{tabular}{ccccc}
  \toprule
    &{meta-$\alpha$}&{meta-$\alpha\&\phi$}\\
              \midrule
  $\alpha$ &0.10 & 0.12\\
    \bottomrule
  \end{tabular}
\end{center}
  \end{minipage}
 \hfill
 \begin{minipage}[t]{0.48 \textwidth}
     \caption{Learned value of $\alpha$ of meta-$\alpha$ and meta-$\alpha\&\phi$ on MNIST.}
     \label{tab:mnist-a}
\begin{center}
  \begin{tabular}{ccccc}
  \toprule
    &meta-$\alpha$ &meta-$\alpha\&\phi$ \\
              \midrule 
  $\alpha$ &0.14  &0.80 \\
    \bottomrule
  \end{tabular}
\end{center}
  \end{minipage}
\end{table}

\subsection{Image Generation with Variational Auto-Encoders}

During meta-training, we sample 128 images of each task/digit and use half of the images to compute Eq.(\ref{eq:phi-update-d}) and the other half for computing the meta-loss. The number of training epochs is 600. We set $K=10$. For all methods, we use the same architecture (100 hidden units and 3 latent variables) and the marginal log-likelihood estimator as in \citet{kingma2013auto}. We report the learned value of $\alpha$ from meta-$\alpha$ and meta-$\alpha\&\phi$ in Table~\ref{tab:mnist-a}. \rzz{The meta-loss is the negative marginal log-likelihood.}

\subsubsection{Examples of Reconstructed Images}
\begin{figure}[h!]
    \centering
    \begin{tabular}{cccc}
    \includegraphics[width=3.cm]{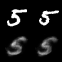}&
    	\includegraphics[width=3.cm]{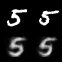} &
    	\includegraphics[width=3.cm]{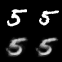} 
    	\\
    	(a) VI$\&\phi$ &(b) meta-$\alpha\&\phi$ & (c) meta-$f\&\phi$
    \end{tabular}
    \caption{Reconstructed images of digit 5 by Meta-$D\&\phi$.}
    \label{fig:minist-sample}
\end{figure}

\rzz{Besides comparing marginal likelihood in Section \ref{sec:exp:mnist}, we visualize the reconstructed images on two examples of digit 5. As shown in Fig \ref{fig:minist-sample}, our methods produce higher quality of images because the reconstructed images are sharper and are able to capture different writing styles whereas the images from standard VI are blurry.}

\subsection{Recommender System}
We split the users into seven age groups: under 18, 18-24, 25-34, 35-44, 45-49, 50-55 and above 56, and regard predicting the ratings of users within the same age group as a task since the users with similar age may have similar preferences. We select 4 age groups (under 18, 25-34, 45-49, above 56) as training tasks, and use the remaining as test tasks. \rzz{The meta-loss is the negative log-likelihood as used in \citet{ma2018partial}. }

For meta-$D$ (Algorithm \ref{alg:meta-d}), during meta-training, we sample 100 users per task (400 users in total) and use half of the observed ratings to compute Eq.(\ref{eq:phi-update-d}) and the other half for computing the meta-loss. The number of training epochs is 400. During meta-testing, we use 90\%/10\% training-test split for the three test tasks and train p-VAE with the learned divergence. 

For meta-$D\&\phi$ (Algorithm~\ref{alg:meta-d-phi}), we compare our method with getting a p-VAE model initialization only, which can be regarded as a combination of MAML and p-VAE (denoted VI$\&\phi$). During evaluation, we apply 60\%/40\% training-test split for the test tasks and train the learned p-VAE model with the learned divergence.

\subsubsection{Additional Experimental Results}
\begin{figure}[h!]
    \centering
    \includegraphics[width=5.5cm]{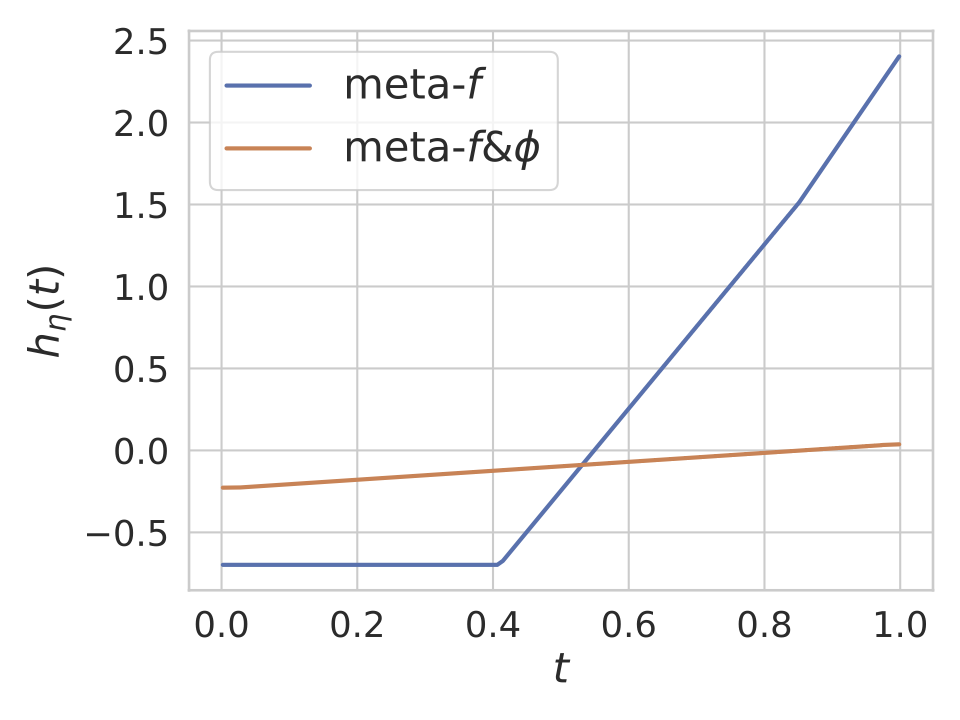}
    \caption{Visualization of learned $h_{\eta}(t)$ from meta-$f$ and meta-$f\&\phi$ on MovieLens.}
    \label{fig:ml-logf2}
\end{figure}

\begin{figure}[h!]
    \centering
    \includegraphics[width=5cm]{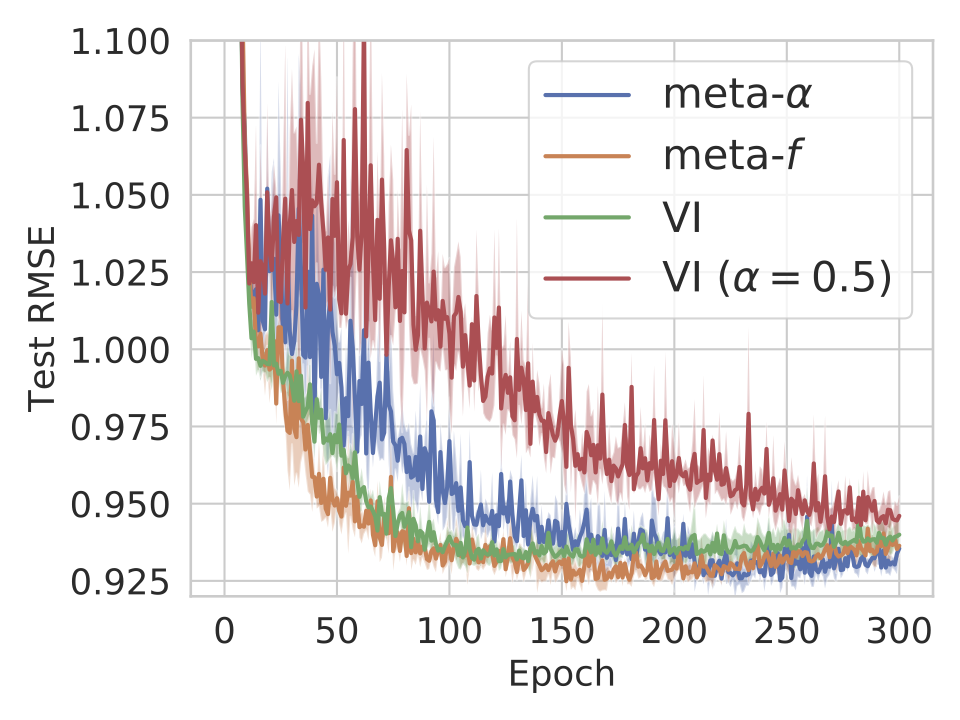}
    \includegraphics[width=5cm]{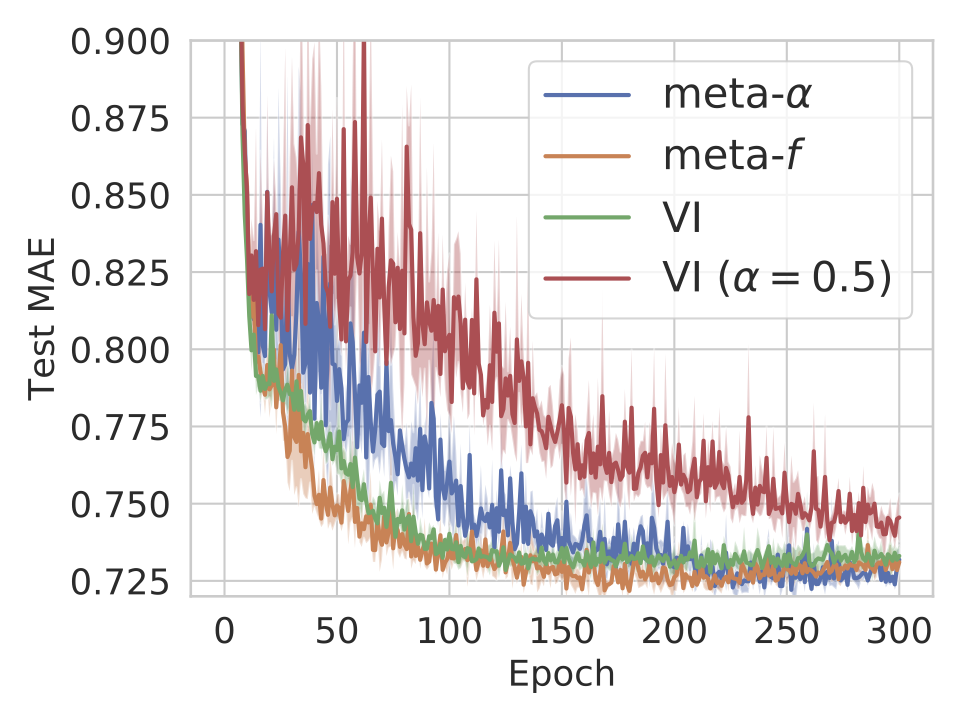}
    \caption{Meta-$D$ on ML: Comparison of meta-$D$ and VI in terms of test RMSE and test MAE.}
    \label{fig:mae-d}
\end{figure}

\begin{figure}[h!]
    \centering
    \includegraphics[width=5cm]{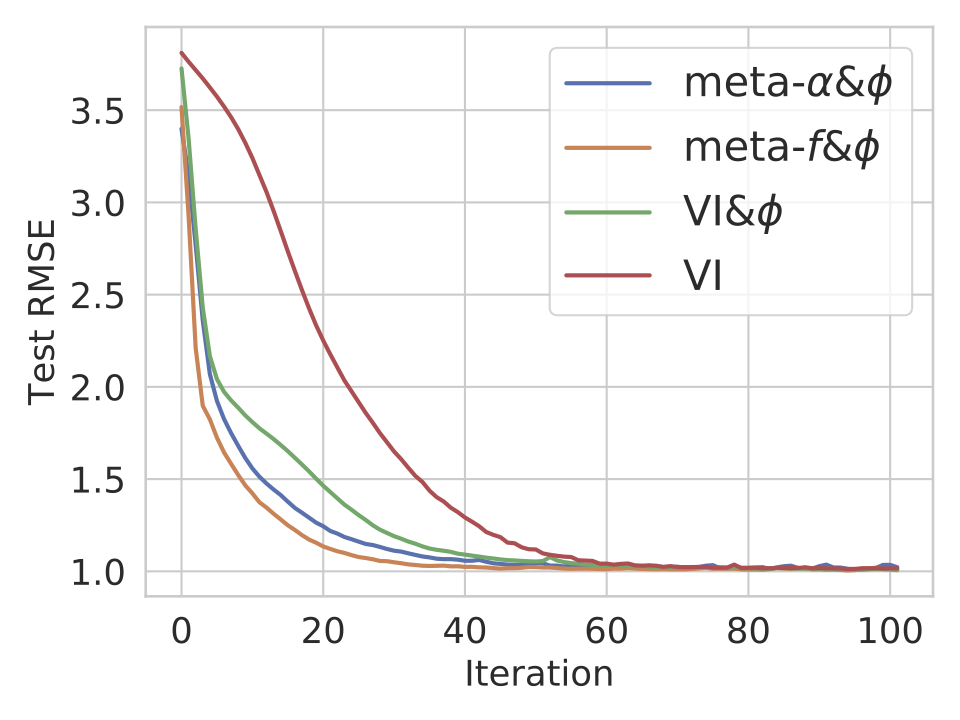}
    \includegraphics[width=5cm]{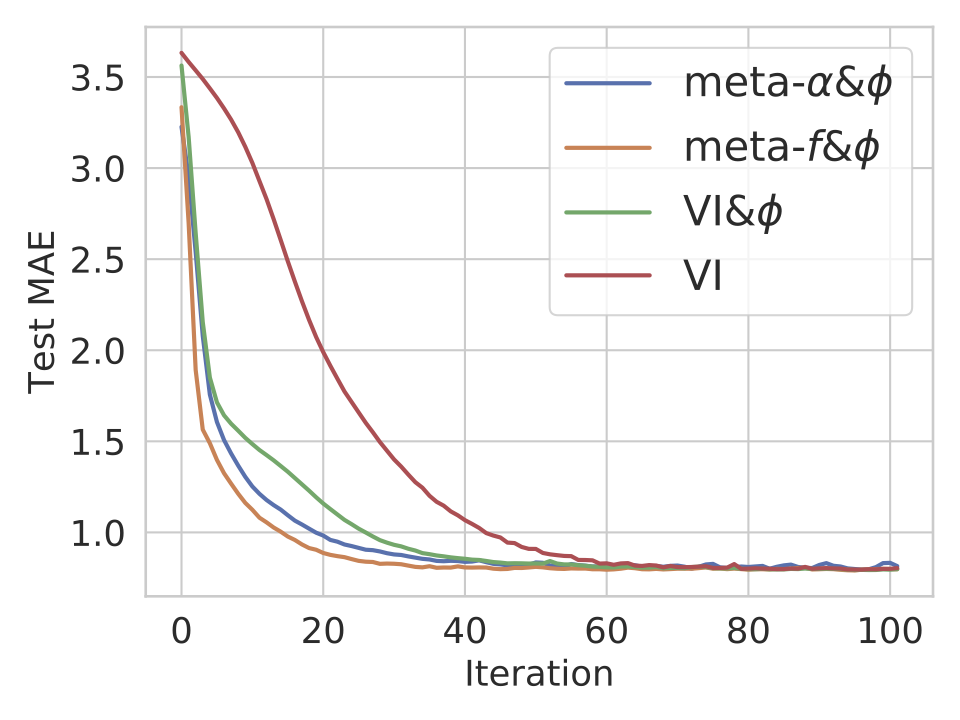}
    \caption{Meta-$D\&\phi$ on ML: Comparison of meta-$D\&\phi$ and VI$\&\phi$ in terms of test RMSE and test MAE.}
    \label{fig:mae-dphi}
\end{figure}

We provide the learned value of $\alpha$ from meta-$\alpha$ and meta-$\alpha\&\phi$ in Table \ref{tab:learned-alpha-ml}. And we visualize the learned $h_{\eta}(t)$ from meta-$f$ and meta-$f\&\phi$ in Figure~\ref{fig:ml-logf2}. Besides the test log-likelihood, there are other popular evaluation metric being used in recommender system and sometimes they are not consistent with each other. Therefore, we also evaluate the performance of our method in terms of other common metrics: test root mean square error (RMSE) and test mean absolute error (MAE). For both metrics, our methods performs better than the baseline in the setting of learning inference algorithm and the setting of learning inference algorithm and model parameters (see Figure \ref{fig:mae-d} and \ref{fig:mae-dphi}).

\begin{table}[H]
  \caption{Learned value of $\alpha$ of meta-$\alpha$ and meta-$\alpha\&\phi$ on MovieLens.}
  
\begin{center}
 \label{tab:learned-alpha-ml}
  \begin{tabular}{ccccc}
  \toprule
    &{meta-$\alpha$}&{meta-$\alpha\&\phi$}\\
              \midrule
  $\alpha$ &0.90& 1.06\\
    \bottomrule 
  \end{tabular}
\end{center}
\end{table}

\rzz{\subsection{Comparison with MLAP \citep{amit2018meta}}

Meta-Learning by Adjusting Priors (MLAP) is a method that meta-learns the Bayesian prior. We compared our method with it to show the importance of learning divergence. We tested on the permuted pixels experiment on MNIST, following the same experimental setup in \citet{amit2018meta}. As this is a few-shot learning setup, we run meta-$\alpha\&\phi$ (meta-learning divergences and variational parameter). Our method attained test error $2.97\%$ which outperformed the best result $3.40\%$ (attained by MLAP-M) in \citet{amit2018meta} significantly. This further demonstrates the importance of learning divergence and the effectiveness of our method on finding the suitable divergence.}

\end{document}